\documentclass[10pt,twocolumn,letterpaper]{article}

\usepackage{iccv}
\usepackage{times}
\usepackage{epsfig}
\usepackage{graphicx}
\usepackage{amsmath}
\usepackage{amssymb} 
\usepackage{color}
\usepackage{booktabs}
\usepackage{etoolbox}

\usepackage[breaklinks=true,bookmarks=false]{hyperref}

\iccvfinalcopy %

\def\iccvPaperID{7572} %

\ificcvfinal\fi

\makeatletter
\apptocmd\@maketitle{{\teaserfigure{}\par}}{}{}
\makeatother

\def\myshift#1{\raisebox{0.5ex}}
\newcommand{\teasernobox}{
    \begin{subfigure}[b]{\linewidth}
        \centering
        \includegraphics[width=\linewidth]{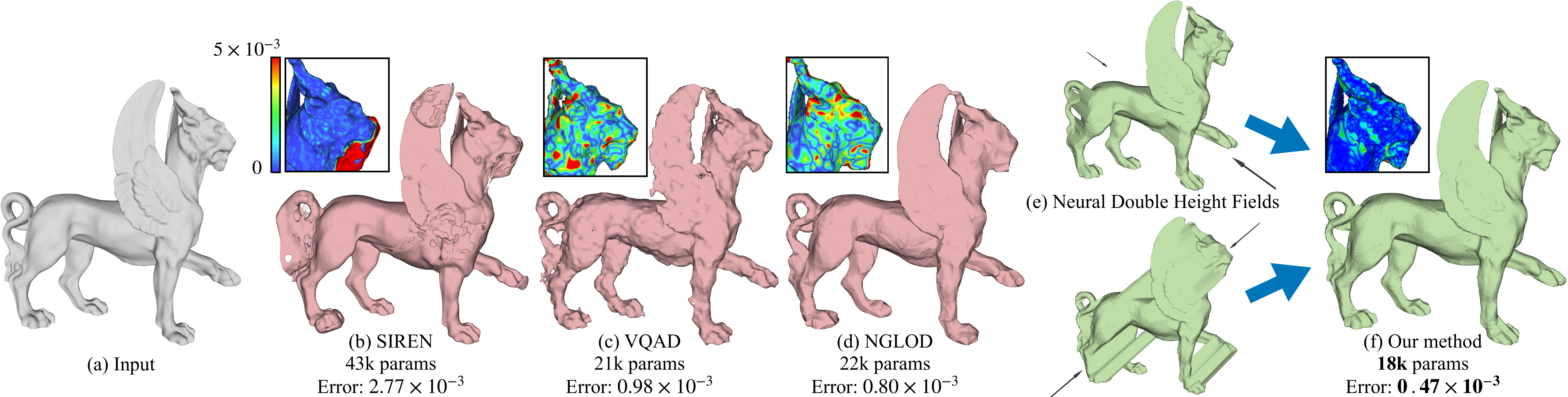}
    \end{subfigure}
}

\newcommand{\teaserfigure}{
    \vspace{-5mm}
    \captionsetup{type=figure}

    \teasernobox

    \vspace{-2mm}
    \setcounter{figure}{0} %
    \captionsetup{type=figure}
    \captionof{figure}{
        \textbf{CN-DHF:} Our CN-DHF method (f) compactly represents complex input shapes (a) using hybrid implicits. We compute DHF surfaces (e) that cover the input and represent them using an implicit 2D neural encoding (arrows indicate DHF axis direction). We define our hybrid implicit model as the intersection of these surfaces. Our hybrid models are both more accurate and more compact than the alternatives (b-d). Insets show approximation error (blue to red color-scheme).%
    }
    \label{fig:teaser}
    \vspace{2em}
}

\vspace{-1.2em}

\usepackage[labelsep=period,skip=1pc,size=small]{caption}
\usepackage{subcaption}
\usepackage{booktabs}

\newcommand{\Fig}[1]{Fig.~\ref{fig:#1}}
\newcommand{\Tab}[1]{Table~\ref{tab:#1}}
\newcommand{\eq}[1]{Eq.~\ref{eq:#1}}
\newcommand{\Sec}[1]{Sec.~\ref{sec:#1}}

\definecolor{turquoise}{cmyk}{0.65,0,0.1,0.3}
\definecolor{purple}{rgb}{0.65,0,0.65}
\definecolor{dark_green}{rgb}{0, 0.5, 0}
\definecolor{orange}{rgb}{0.9, 0.6, 0.1}
\definecolor{red}{rgb}{0.8, 0.2, 0.2}
\definecolor{darkred}{rgb}{0.6, 0.1, 0.05}
\definecolor{blueish}{rgb}{0.0, 0.3, .6}
\definecolor{light_gray}{rgb}{0.7, 0.7, .7}
\definecolor{pink}{rgb}{1, 0, 1}
\definecolor{greyblue}{rgb}{0.25, 0.25, 1}
\definecolor{amethyst}{rgb}{0.6, 0.4, 0.8}

\newcommand{\View}{{\mathcal{V}}}
\newcommand{\numview}{{N}}
\newcommand{\view}{{i}}
\newcommand{\Net}{{\mathcal{F}}}
\newcommand{\netparam}{{\boldsymbol{\theta}}}

\newcommand{\Occupancy}{{\mathcal{O}}}
\newcommand{\Reward}{{\mathcal{R}}}
\newcommand{\Heightfield}{{\mathcal{H}}}
\newcommand{\Lapop}{{\Delta}}
\newcommand{\Rot}{{\mathbf{R}}}

\newcommand{\near}{\text{--}}
\newcommand{\far}{\text{+}}
\newcommand{\dnear}{{h_{\near}}}
\newcommand{\dfar}{{h_{\far}}}

\newcommand{\nearfar}{{\near{/}\far}}
\newcommand{\gtdnearfar}{{\hat{h}_{\nearfar}}}
\newcommand{\dnearfar}{{h_{\nearfar}}}

\newcommand{\IR}{{\mathbb{R}}}
\newcommand{\IE}{{\mathbb{E}}}

\newcommand{\point}{{\mathbf{p}}}
\newcommand{\area}{{\mathbf{a}}}
\newcommand{\earea}{{\tilde{\mathbf{a}}}}
\newcommand{\visible}{{\mathbf{o}}}
\newcommand{\idxtriangle}{{t}}
\newcommand{\loss}[1]{{\mathcal{L}_{\text{#1}}}}
\newcommand{\Lone}{{L_1}}

\newcommand{\supp}{supplementary material\xspace}
\newcommand{\mylesdata}{Myles \etal~\cite{Myles16}\xspace}

\newcommand{\customparskip}{0.5em}
\renewcommand{\paragraph}[1]{\vspace{\customparskip}\noindent\textbf{#1}}

\usepackage{multicol}

\usepackage{longtable}  %

\usepackage{lipsum}

\begin{document}

\title{CN-DHF: Compact Neural Double Height-Field Representations of 3D Shapes}

\author{Eric Hedlin\\
University of British Columbia\\
{ iamerich@cs.ubc.ca}
\and
Jinfan Yang\\
University of British Columbia\\
{\tt\small  yangjf@cs.ubc.ca}
\and
Nicholas Vining\\
NVIDIA\\
{\tt\small  nvining@cs.ubc.ca}
\and
Kwang Moo Yi\\
University of British Columbia\\
{\tt\small  kmyi@cs.ubc.ca}
\and
Alla Sheffer\\
University of British Columbia\\
{\tt\small  sheffa@cs.ubc.ca}}

\maketitle

\ificcvfinal\thispagestyle{empty}\fi

\begin{abstract}

    We introduce {\em CN-DHF (Compact Neural Double-Height-Field)}, a  novel hybrid neural implicit 3D shape representation that is dramatically more compact than the current state of the art.
    Our representation leverages Double-Height-Field (DHF) geometries, defined as closed shapes bounded by a pair of oppositely oriented height-fields that share a common axis, and leverages the following key observations: DHFs can be compactly encoded as 2D neural implicits that capture the maximal and minimal heights along the DHF axis;  and typical closed 3D shapes are well represented as intersections of a very small number (three or fewer) of DHFs.
    We represent input geometries as CN-DHFs by first computing the set of DHFs whose intersection well approximates each input shape, and then encoding these DHFs via neural fields.
    Our approach delivers high-quality reconstructions, and reduces the reconstruction error by a factor of $2.5$ on average compared to the state of the art, given the same parameter count or storage capacity.  Compared to the best performing alternative, our method produced higher accuracy models on 94\% of the 400 input shape and parameter count combinations tested.
\end{abstract}
\section{Introduction}
\label{sec:intro}

Finding accurate compressed representations of 3D shapes is an important active research area, with applications including video games, 3D content streaming, reinforcement learning, and VR/AR environments.
Neural shape representations ~\cite{nglod,instantngp,vqad,siren} provide a promising avenue for addressing this problem, can accurately encode highly detailed shapes using just a few thousand parameters, and have a vast array of additional potential applications including generative models of shape spaces~\cite{deepsdf}, neural rendering~\cite{volsdf,neus,unisurf}, and Simultaneous Localization and Mapping~(SLAM)~\cite{li2023dense}.
These works build upon deep network based shape representations~\cite{deepsdf, occnet, bspnet,ChenSDFs}, and leverage advances in neural field research~\cite{vincentsurvey}.
Implicit neural shape representations support instantaneous in-out queries, and are particularly well suited for ray-tracing based rendering and generative modeling~\cite{deepsdf,nglod}.
Related approaches learn both shape and appearance together for highly detailed novel-view synthesis~\cite{volsdf,neus,unisurf}.
In all of these works, a core challenge is representing the original 3D shape as accurately as possible, while keeping the model size, or parameter count low.
Parameter count directly impacts the amount of storage required by the models, as well as transmission, rendering and processing times.
We present a new implicit {\em hybrid} neural representation of 3D shapes that dramatically improves the accuracy-versus-size trade-off compared to the state of the art.%

\begin{figure}
    \centering
    \includegraphics[width=\linewidth]{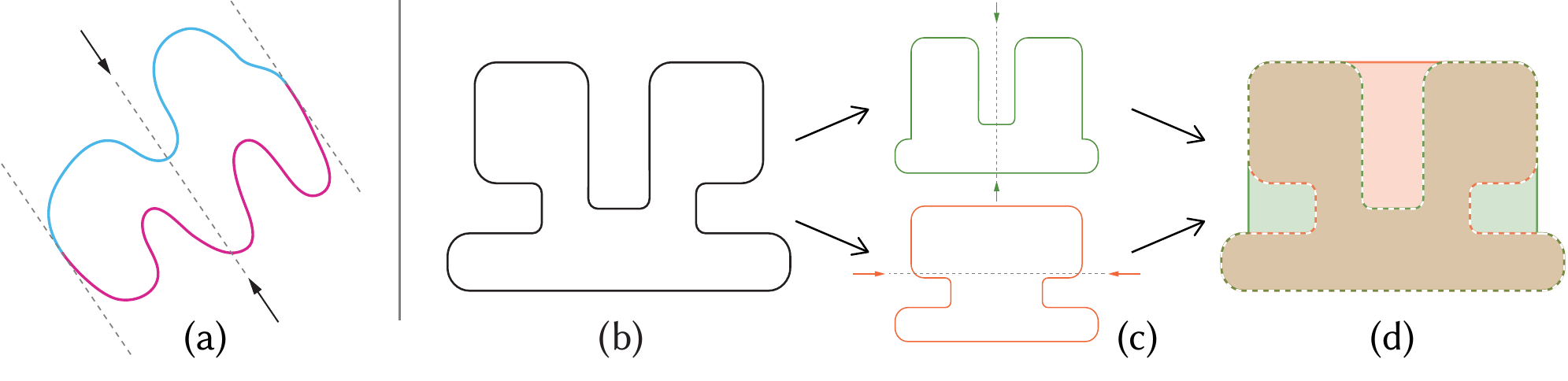}
    \caption{{\bf DHF Surfaces}: (a) a DHF surface (axis shown by black arrows, top height-filed surface in blue, bottom in purple). General shapes (b) can typically be accurately described via an intersection of a small number of DHF surfaces (c,d); in (c) arrows indicate DHF axes directions. }
    \label{fig:dhf}
\end{figure}

Our method is inspired by state-of-the-art hybrid shape representation approaches that compactly and accurately represent complex 3D content by combining neural networks with traditional 3D shape representations, such as octrees~\cite{nglod,instantngp} or displacement maps ~\cite{idf}.
We posit that one can achieve even higher representation compactness via a simple yet powerful observation that has been well tested
over time---lower dimensional content can be represented more compactly.
We consequently represent 3D geometries as 2D neural fields.
Historically, 2D representation of 3D shapes has been achieved via parametric, rather than implicit, representations such as Geometry Image atlases~\cite{sheffer2007mesh,geometryimages,sander2003multi,groueix2018}.
While fairly compact, such parametric approaches require complex manipulation to evaluate in/out queries or compute ray-surface intersections.

We represent 3D shapes via an {\em implicit} encoding that uses 2D neural fields.
Key to our representation are Double-Height-Field (DHF) surfaces: closed surfaces formed by a union of two height-fields defined with respect to two opposite directions of the same axis (\Fig{dhf}a)~\cite{yang2020dhfslicer}.
We observe that DHF surfaces can be represented as implicit shapes bounded by their top and bottom surfaces; allow for quick in/out queries, facilitating ray-tracing and other applications; and can be stored as two channel (top and bottom heights) images.
More importantly, DHFs can be compactly and accurately represented by storing their DHF axis and encoding their surfaces as 2D neural fields that capture the maximal and minimal heights along the axis (\Sec{method_neural}).
We use these neural DHF surfaces as the core building block of our shape representation.

A potential approach to extend this representation to general non-DHF shapes (\Fig{dhf}b) is to segment them into DHF blocks \cite{yang2020dhfslicer,alderighi2021volume}.
However using a union of non-overlapping volumetric blocks as a shape representation can lead to geometry mismatches along internal boundaries; such mismatches can trigger catastrophic in-out query failures (as points deep inside the original shape can be just outside all union blocks).
We avoid such instabilities by observing that the vast majority of shapes used in virtual environments can be well represented via {\em intersection} of a small number of DHF surfaces (\Fig{dhf}cd).
Notably an intersection of implicit surfaces is implicit by construction and allows for robust in-out query evaluation in the object interior.
Our analysis of representative datasets suggests that the vast majority of shapes can be well approximated by intersecting three DHFs or less; and approximately half the shapes analyzed are well captured by single DHFs (Sec.~\ref{sec:results}).
Due to the small number of intersected surfaces, shapes defined via DHF intersection allow for real-time processing.

We generate our desired {\em Compact Neural DHF}, or {\em CN-DHF}, representations via a hybrid framework that consists of a geometric algorithm for finding the DHFs whose intersection well-approximates our input surface, and a Multi-Layer Perceptron (MLP) for modeling the individual DHFs.
We first find the DHF surface that contains the input shape and best approximates it (\Fig{dhf}c,top), and then proceed to compute DHF surfaces that contain the input and best approximate the portions of the input surface not covered by previously computed DHFs (\Fig{dhf}c,bottom).
We model each DHF using the SIREN architecture~\cite{siren}, encoding both high- and low-frequency details.
Our loss function combines a positional term aiming to accurately predict the maximal and minimal heights along a DHF axis, and a Laplacian term that captures the local features of the top and bottom DHF surfaces (\Sec{neuralfield}).

We demonstrate our method and validate its effectiveness by learning CN-DHF representations of 100 diverse shapes from representative shape repositories~\cite{abcdataset,thingi10k,yang2020dhfslicer,StanfordScanRep}, using four different parameter counts each.
We compare our results to those generated by three leading alternatives~\cite{nglod,vqad,siren}.
94\% of our learned models more accurately approximate the input ground truth shapes than those produced by the best performing  alternative method using same or larger parameter counts.
On average our outputs are $2.5$ times as accurate as those produced by the best alternative method with the same model parameter budget.

To summarize, our contributions are threefold:
(1) we propose CN-DHF, a new implicit hybrid neural representation that encodes 3D shapes using a small set of 2D images (DHFs);
(2) we show that CN-DHF models of 3D shapes are highly accurate and, for the same storage capacity,
reduce the reconstruction error by a factor two compared to SOTA alternatives;
(3) finally, we show that most surfaces used to populate virtual environments can be accurately represented using an intersection of three or fewer DHFs.

\section{Related Work}
\label{sec:related}

Our work builds on both traditional and neural network-based approaches to compact 3D shape representation, and on methods for single and double height field processing.

\paragraph{Meshes.}
While triangle and quadrilateral meshes remain the de facto standards for representing 3D shapes for various Graphics applications, they are known to be inherently wasteful in terms of their memory footprint~\cite{geometryimages} and are known to be challenging to learn on, due to topological irregularity \cite{MeshCNN,maron}; compressed meshes \cite{toumagotsman,peng2005technologies} have a smaller footprint but are harder to manipulate.

\paragraph{Geometry Image Atlases.}
3D surfaces can be represented in parametric form via a 2D-to-3D map. The seminal Geometry Images method~\cite{geometryimages} leveraged this property to effectively compress surfaces via 2D parameterization: input meshes are cut into topological discs, which are parameterized into the plane, and the 3D coordinates of points on the disc are stored in image form.
Follow-ups~\cite{sander2003multi,Carr:2006:RMG} reduce parametric distortion and improve reconstruction accuracy by supporting atlases with multiple disc charts.

2D-to-3D parameterization is exploited by methods such as AtlasNet \cite{groueix2018} and its followups \cite{deprelle2019learning,deprelle2022learning} for learning shape spaces.
While using neural networks for atlases allows compact encoding,
they are inherently ill suited for tasks such as in-out queries or ray-tracing as the inverse mapping from 3D to 2D is non-trivial.
We complement these approaches by introducing an implicit compact neural representation based on 2D encoding.
Our representation is better suited for many applications that target implicits, as discussed below.

\paragraph{Implicit Representations.}
Implicit, or level set, representations of closed surfaces~\cite{BlinnImplicits, osher2004level,wyvill1998blob, schmidt2005sketch} have a long history in graphics and vision literature. Implicit representations are well suited for in-out queries, ray-tracing and various applications~\cite{jones20063d,Takikawa2022SDF} including digital humans \cite{saito2019}, inverse rendering \cite{zhang2021} and robotics perception\cite{zhu2021rgb}.
Implicit representations popular in learning settings include occupancy maps \cite{occnet} and Signed Distance Functions~(SDF)~\cite{ChenSDFs}.
While analytic implicits are highly compact, converting generic surfaces into analytic implicit form remains an open problem \cite{reverseengsurvey}; the more commonly used grid-based representation of implicits (e.g. \cite{musethvdb}) is highly memory consuming.

Consequently, significant effort has been employed to reduce their space footprint.  This includes using neural networks to learn compact functions~\cite{deepsdf, occnet, imnet,sitzmann2019metasdf}, or, more traditionally, focusing on adaptive multiresolution hierarchies~\cite{LH07}.
More recently, Neural Geometric Level of Detail (NGLOD)~\cite{nglod} combines both research directions into one by using a sparse hierarchical grid together with neural networks.
Variable bitrate neural fields (VQAD)~\cite{vqad} further compresses representations by using vector quantization. While more efficient than naive storage, these methods still remain in the 3D realm.
CN-DHFs achieve much higher compactness by reducing the dimensionality of the neural field, while retaining the desirable features of traditional occupancy maps that make them well suited for numerous downstream applications.

Yifan et al.~\cite{idf} propose combining a SDF base surface with a high-frequency displacement map. Their method relies on the existence of clean geometry with well-defined, smooth normals.  Our method makes no such assumptions, and robustly processes inputs which do not satisfy these properties (Sec.~\ref{sec:results}); moreover, as our comparisons show, CN-DHF achieves higher accuracy at one tenth of the model size compared to their method.

\paragraph{Single and Double  Height-Fields.}
Height-field surfaces are explicit surfaces that can be bijectively parameterized via projection to a plane orthogonal to their axis (Fig~\ref{fig:dhf} blue and purple). Segmenting surfaces into height-field charts and parameterizing these charts in 2D can be seen as a special case of Geometry Image parameterization; however such parameterization can exhibit high distortion and consequently low accuracy, and requires a high chart count ~\cite{FeketeMiller,Hu, Alemanno2014, herholz2015approximating, gao2015revomaker, Muntoni2018, Muntoni2019}. Novak and Dachsbacher~\cite{novak2012rasterized} represent triangle meshes via bounding volume hierarches where each leaf in the hierarchy is a single height field. While suitable for ray tracing this representation is neither compact nor implicit.  DHF surfaces were introduced by Yang et al.~\cite{yang2020dhfslicer} in the context of CNC milling. Both Yang et al.~\cite{yang2020dhfslicer} and Alderighi et al.~\cite{alderighi2021volume} propose methods for decomposing 3D shapes into DHF blocks. Using a union of non-overlapping blocks as a shape representation can lead to misalignments along internal boundaries and thus can trigger in-out query failures even deep inside the original shape, since points next to interior boundaries can end up outside all union blocks.

We are the first to introduce a neural compact DHF representation that leverages their unique properties, and the first to recognize that DHFs form a class of implicit surfaces and define them accordingly.
Rather than a union of blocks, we represent general shapes as \emph{intersections} of closed DHF surfaces.
Using intersection rather than union speeds-up negative in-out queries, and sidesteps the need to handle any gap artifacts along inter-block boundaries.

\section{Method}
\label{sec:overview}

CN-DHF represents each shape as a collection of neural double height fields with distinct DHF axes; the shape is then reconstructed by taking the intersection of the height fields (\Sec{inference}).
Importantly, we do not learn the axes used for each DHF, or the number of DHFs required; rather, we optimize the choice and number of axes for maximum coverage (\Sec{view}). Once the DHFs are known, the learning problem is reduced to fitting a neural field to each height field (\Sec{neuralfield}) We detail each step below.

\begin{figure}
    \includegraphics[width=\linewidth]{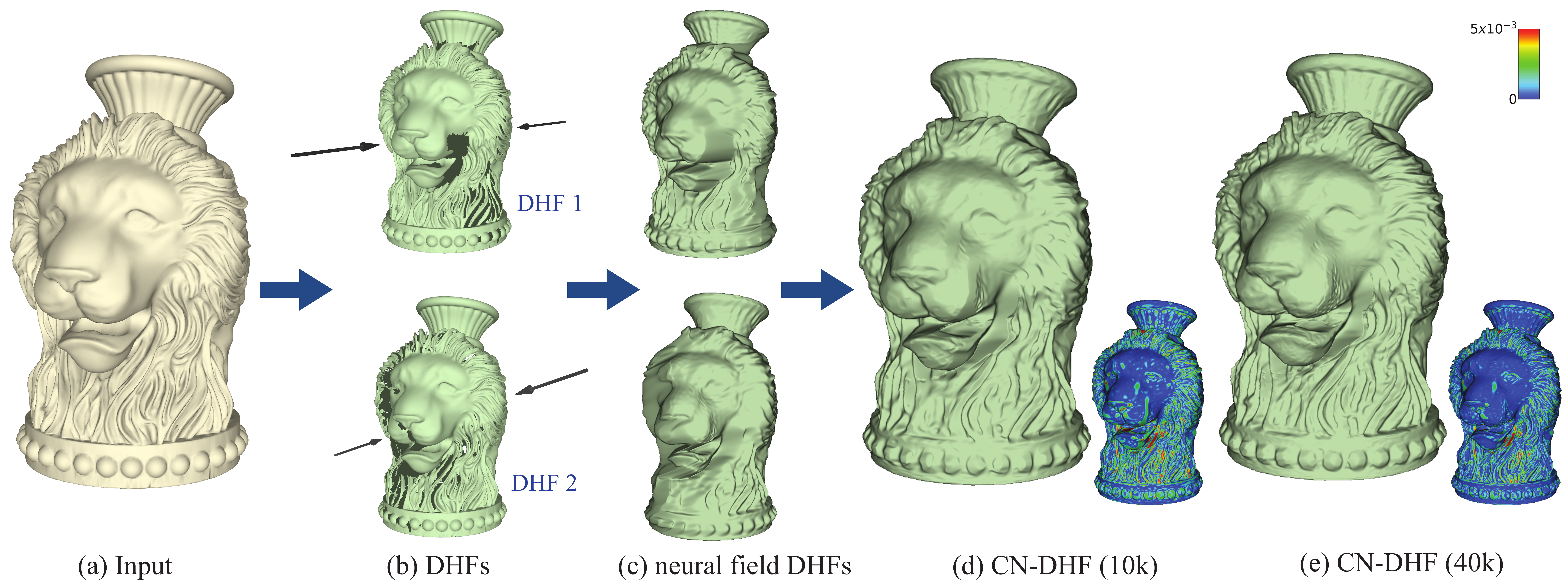}
    \caption{{\bf CN-DHF overview:} We identify the DHFs (b) that jointly cover the input shape (a) and learn neural field implicit models of each (c, 5K parameters each), our CN-DHF model is defined by their intersection (d).  Model accuracy (shown by the error heatmap insets) improves with increase in parameter count~(e).
    }
    \label{fig:overview}
\end{figure}

\subsection{Occupancy}
\label{sec:inference}

We start by reviewing how a single DHF is defined mathematically~\cite{yang2020dhfslicer}.
Given a single DHF in a coordinate system where the $z$ axis coincides with the DHF axis (the normal of the DHF plane), for each 2D point $(x,y)$ the DHF shape is defined via its maximal and minimal $z$-values $z_{\far}(x,y)$ and $z_{\near}(x,y)$.
For $(x,y)$ outside the shape, we set $z_{\far}(x,y) = -1, z_{\near}(x,y)=1$ by convention.

We observe that this definition can be seen as an implicit definition of a shape, and that closed DHFs constitute a sub-class of implicit surfaces.
Specifically, given a DHF surface, for each point $(x,y,z)$ the point is on the surface if $z$ is equal to $z_{\far}(x,y)$ or $z_{\near}(x,y)$ and $z_{\far}(x,y)>z_{\near}(x,y)$; is inside if  $z_{\far}(x,y)>z> z_{\near}(x,y)$; and is outside otherwise.
We can encode this DHF as a neural field $\Net_z$ that, for a given point $(x,y)$, returns the DHF values $z_{\near}(x,y)$ and $z_{\far}(x,y)$. The occupancy for a given point $\point=(x,y,z)$ can then be written as
\begin{equation}
    \begin{aligned}
        \Occupancy_z(\point) =
        \begin{cases}
            1, & \text{if } z_{\near}(x,y) \leq z \leq z_{\far}(x,y) \\
            0, & \text{otherwise}
        \end{cases}
    \end{aligned}
\end{equation}

We can generalize this to a DHF with an arbitrary axis $\View$ by rotating $\point$ to $\Rot\point$, where $\Rot$ is the minimal rotation that transforms $\View$ to the $z$ axis, and where the remaining two axes are chosen consistently for a given $\View$. If we encode this DHF as a neural field $\Net_\View$ that, for a given point $(u,v)$, returns the DHF values $\dnear(u,v)$ and $\dfar(u,v)$, the occupancy for a given point $\point=(x,y,z)$ can be written as
\begin{equation}
    \begin{aligned}
        \Occupancy_{\View}(\point) =
        \begin{cases}
            1, & \text{if } \dnear(\Rot\point_x,\Rot\point_y) \leq \Rot\point_z \leq \dfar(\Rot\point_x,\Rot\point_y) \\
            0, & \text{otherwise}
        \end{cases}
    \end{aligned}
    \label{eq:per_view_occupancy}
\end{equation}
Finally, given a set of DHFs with axial directions $\left\{\View_1, \View_2, \dots, \View_\numview\right\}$ for a shape, where $\View_\view \in \IR^3$, and associating each axis $\View_\view$ with the neural field $\Net_\view$, we may write the occupancy for the shape as:
\begin{equation}
    \begin{aligned}
        \Occupancy(p) = \min_{\view}\Occupancy_{\View_\view}(p)
    \end{aligned}
    \label{eq:occupancy}
\end{equation}

Note that the $\min$ operation here can be seen as a `logical and' operation. At this point, we could treat $\Occupancy(p)$ as a typical occupancy function learned directly via a neural network; however, unlike previous work, we are creating 3D representation out of information stored as 2D planes which is much more efficient as we show in \Sec{results}.

\subsection{Optimal DHF Decomposition}
\label{sec:view}
To find an optimal representation of a shape as a minimal set of DHFs with axes $\left\{\View_1, \View_2, \dots, \View_\numview\right\}$, we maximize the area of the surface triangles that are represented either by some $\dnear$ or $\dfar$. We may think of this as being`visible' when viewing the surface from a camera point along the selected DHF axis; this corresponds to the notion of accessibility in Yang et al.~\cite{yang2020dhfslicer}. We first discuss how we define visibility in a computationally tractable way, and subsequently how we find the best axes according to this definition.

\paragraph{Visibility.} To keep our computation tractable, we define visibility via randomly sampling points on the surface, and tracing along the DHF axis in both directions. A point on a triangle is said to be {\em occluded} if both rays along the DHF axis direction and emanating at the point on the surface intersect another triangle, and {\em visible} otherwise. For each triangle, if the triangle contains {\em any} points that are occluded, we consider the entire triangle to be occluded; we define occlusion of triangles this way as we rely on random surface points which do not guarantee full coverage of the triangle. The amount of visible area on the input surface for a view $\View_\view$ is therefore the area of all non-occluded triangles.

We sample a total of $N_p{=}5\cdot N_t$ points on the input surface, where $N_t$ is the number of triangles on the input surface, and distribute them evenly by surface area. If $A$ is the total surface area of the input mesh, we aim to sample $n_p{=}\frac{A}{N_p}$ points per unit of area; for a triangle $t$ on the input mesh with area $\area_t$, we therefore sample $n_{p_t}{=}\lceil n_p\cdot\area_t \rceil $ points on its surface. We always sample at least one point on each triangle, and the centroid of the triangle is always included in the sampled points; the remaining points are sampled uniformly in barycentric space on the triangle. We write the visibility of a triangle $\idxtriangle$ for a given viewing direction $\View_\view$ as:
\begin{equation}
    \visible_\idxtriangle\left(\View_\view\right) =
    \begin{cases}
        1, & \text{not occluded from} \;\; \View_\view \\
        0, & \text{otherwise}
    \end{cases}
\end{equation}

\paragraph{Finding DHF Axes.} We find the best set of DHF axes by repeatedly evaluating the visible surface area from multiple candidate axes. We sample fifty candidate directions on the unit sphere using spherical Fibonacci sampling~\cite{Keinert:2015}, and three additional axes corresponding to the three major axes as many objects encountered in practice are axis-aligned~\cite{canonicalcapsules,fu2008upright}.
For each axis, we consider the following criteria. First, triangles that are not visible from any axis can be ignored, as we are interested in the outer surface only. Second, some triangles may only be visible from a few axes; in order to minimize the total number of DHF directions used to represent the input shape, we ideally wish to use these directions first if possible.

Thus, denoting the area of a triangle as $\area_\idxtriangle$, we write the `effective' surface area of a triangle $\earea_\idxtriangle$ as:
\begin{equation}
    \earea_\idxtriangle =
    \area_\idxtriangle \max\left(
    \frac{\sum_\view\visible_\idxtriangle(\View_\view)}{\IE_{\idxtriangle}\left[\sum_\view\visible_\idxtriangle(\View_\view)\right]}, 1
    \right)
\end{equation}
Since we are aiming to maximize the area given multiple axes, the reward $\Reward$ we seek to maximize can be written as:
\begin{equation}
    \Reward\left(\left\{\View_1, \View_2, \dots, \View_\numview\right\}\right)
    =
    \sum_\idxtriangle \earea_\idxtriangle \max_{\view} \left[
        \visible_\idxtriangle\left(\View_\view\right)
        \right]
\end{equation}
Maximizing this reward in an optimal sense, either continuously or with discrete randomly sampled axes, is computationally prohibitive.
We opt for a greedy solution where we iteratively choose the axis that maximizes the reward growing the number of DHF axes. After an axis is selected, we remove all triangles visible from that axis from consideration in subsequent rounds. We stop this process when 99.9\% of the effective surface area $\sum_{t} \earea_t$ is covered.

\subsection{Per-axis Neural Fields}
\label{sec:neuralfield}
\label{sec:method_neural}

We now describe the per-axis neural fields $\Net_{\View_\view}$ and how we train them.
For the remainder of this subsection, we drop $\view$ for simplicity.
Formally, for each axis direction $\View$, we associate it with a neural field encoding the double height field values $[\dnear, \dfar]=\Net_{\View}\left(u, v;\netparam\right)$, where $(u,v)$ are the coordinates defining the location on the plane that goes through the origin (shape centre) and has the normal defined by $\View$, $\dnear$ and $\dfar$ are the \emph{signed} distances to the nearest and furthest surfaces from said plane, and $\netparam$ is the set of parameters defining this neural field.

To implement $\Net_\View$ we rely on a SIREN network~\cite{siren}; specifically, we use an Multi-Layer Perceptron (MLP) composed of 5 hidden layers, each of which is activated by a sine function.
The number of neurons for each layer is set to various values ranging from 17 to 130, according to the desired model capacity.
As briefly discussed in \Sec{inference}, the network directly takes $(u,v)$ parametric coordinates on the plane whose normal is the DHF axis and that passes through the origin as input, and outputs the DHF values $\dnear(u,v)$ and $\dfar(u,v)$.
As the final occupancy in \eq{occupancy} is constructed by taking the minimum of the occupancy of these per-axis neural fields, we can simply train each axis independently.
We train $\Net_\View$ according to two objectives: $\loss{reg}$, where the network is trained to regress to accurate height field values, and $\loss{lap}$ so that the \emph{numerical Laplacian} of the estimated height field matches the ground truth which we found to be essential in generating smooth surfaces.
The final loss for each network is thus:
\begin{equation}
    \loss{} = \loss{reg} + \lambda \loss{lap}
\end{equation}
where $\lambda$ is a parameter balancing the two loss terms.
We empirically set $\lambda{=}10$ for all our experiments.
We detail the two loss terms below.

\paragraph{Regressing the height field -- $\loss{reg}$.}
To accurately regress the ground-truth height field values we optimize an $\ell_1$ loss. For a given axis direction $\View$ and UV coordinates $(u,v)$, we find ground truth values $\gtdnearfar$ by raycasting.
One caveat with naively learning to regress the height field values is that the ground-truth height field values do not exist for $(u,v)$ points where the ray along the axis direction does not intersect with any triangle; in other words, there's no target value to regress.
Nonetheless, our formulation in \eq{per_view_occupancy} allows for a straightforward solution to this problem.
As long as
$\dfar < \dnear$,
$\Occupancy_\View(\point)=0$  regardless of the regressed value.
We thus write:
\begin{equation}
    \loss{reg} =
    \begin{cases}
        \left|\gtdnearfar - \dnearfar\right|,    & \text{if} \quad \exists \gtdnearfar \\
        \max\left(0, \dfar - \dnear + C \right), & \text{otherwise}
    \end{cases}
    \label{eq:lossreg}
\end{equation}
where $C{=}0.1$ is a small positive constant set empirically to encourage a small margin between the two orderings.%
Please see the supplemental material for additional details.

\paragraph{Enforcing smoothness -- $\loss{lap}$.}
As shown in \Sec{results}, $\loss{reg}$ alone provides noisy surfaces, which is not the case for most shapes that we wish to model.
We thus regularize that the \emph{numerical Laplacian} matches between the ground truth and our regressed values.
This can easily be implemented, as our representation \emph{by construction} operates on the 2D $(u,v)$ space which is then converted into 3D via height fields.
Denoting the numerical Laplacian operator as $\Lapop$ (which can be implemented easily as a convolution operation in the $(u,v)$ space), the height field images created by $\Net_\View$ as $\Heightfield_{\nearfar}$, and the corresponding ground-truth as $\hat{\Heightfield}_{\nearfar}$, we write:
\begin{equation}
    \loss{lap}=\left|
    \Lapop\Heightfield_{\nearfar} - \Lapop\hat{\Heightfield}_{\nearfar}
    \right|
    \;.
\end{equation}
We only apply this regularization for points where $\hat{\Heightfield}$ is valid; that is, where ground-truth height field values exist.

\section{Results}
\label{sec:results}

\subsection{Experimental Setup}
\label{sec:exp_setup}

\vspace{-\customparskip}
\paragraph{Dataset.}
We evaluate our method on 100 shapes from multiple publicly available data sources to maximize variability (since these sources have a degree of overlap, some inputs are present in several). We sample 40 shapes from Thingi10k~\cite{thingi10k}, including the Thingi32 subset of 32 inputs used by NGLOD~\cite{nglod}; this ensures comparison with shapes chosen by the baselines.
We use 40 inputs from the ABC~\cite{abcdataset} dataset of CAD models (as this dataset contains many basic shapes, such as cylinders, that might bias evaluation towards our method, we curate the subset to focus on complicated shapes).
We use 17 inputs from the DHFSlicer~\cite{yang2020dhfslicer} paper that first introduced the notion of DHF surfaces and experimented with decomposing diverse shapes into DHF blocks.
Finally, we include 3 canonical, complex scanned shapes from the Stanford 3D Scanning Repository \cite{StanfordScanRep}: the {\em buddha}, {\em Thai statue}, and {\em David} models.
The set includes 57 closed manifold meshess, 7 non-manifold meshes, 40 inputs with more than one connected component, and 33 inputs with multiple boundary loops.
\Sec{exp_ablation} includes additional statistics.

\paragraph{Baselines.}
We compare our method against three representative baseline methods.
We compare our results to those of Neural Geometric Level of Detail~(NGLOD)~\cite{nglod} and Variable Bitrate Neural Fields~(VQAD)~\cite{vqad}, two state-of-the-art methods that focus on compact representation of 3D shapes.
As their comparisons show, these methods outperform earlier alternatives such as \cite{instantngp,davies2020overfit} in terms of quality/compactness tradeoff.
We also compare our method against SIREN~\cite{siren}. 
Since SIREN is the base architecture for our neural field, comparisons against it provide an important baseline and demonstrate the impact of moving from encoding a 3D Signed Distance Field (SDF) representation to our 2D CN-DHF representation.
We use default hyperparameter settings for all methods. For SIREN, to make the supervision signal the same for all methods, we disable normal supervision.

While our framework and SIREN allow parameter counts to be controlled directly, NGLOD and VQAD set parameter counts dynamically and only allow level of detail control.
We thus train NGLOD and VQAD at LOD 1 through 4, and set both ours and the SIREN parameter count to roughly match their per-LOD averages (5k, 10k, 20k, and 40k).
We train all methods to one million epochs to ensure convergence. Lastly we compare  to Wang~\etal~\cite{idf} on the subset of inputs, reported in their paper.

\subsection{Comparisons.}
\label{sec:exp}
\label{sec:exp_quant}

\begin{figure*}
    \centering
    \includegraphics[width=\linewidth, trim=90 370 240 230, clip]{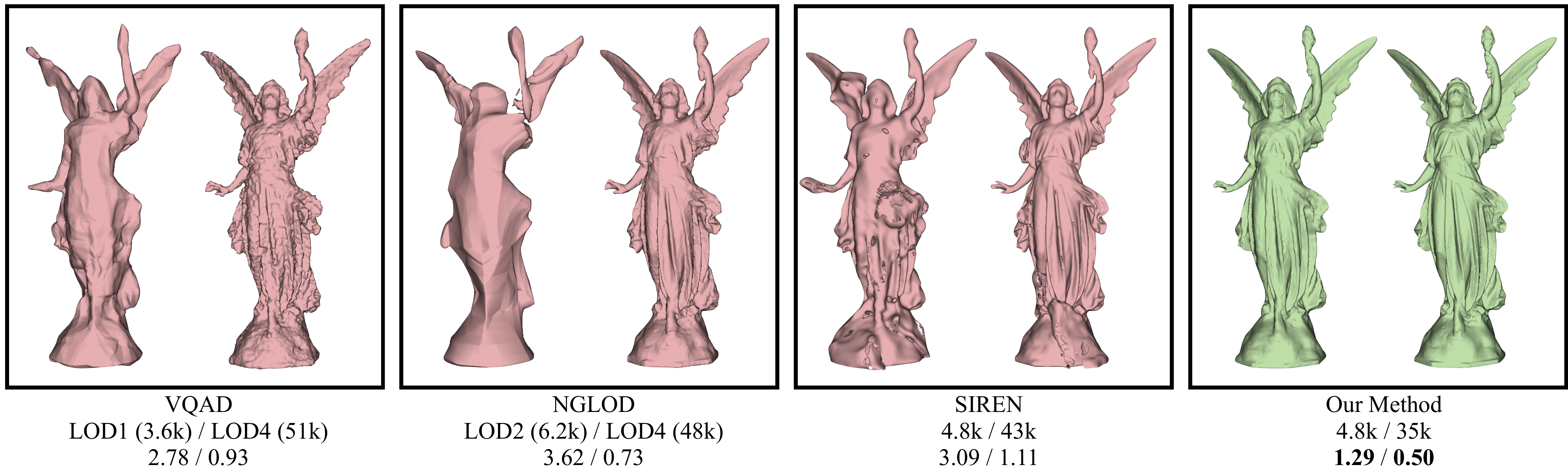}
    \includegraphics[width=\linewidth, trim=90 370 240 230, clip]{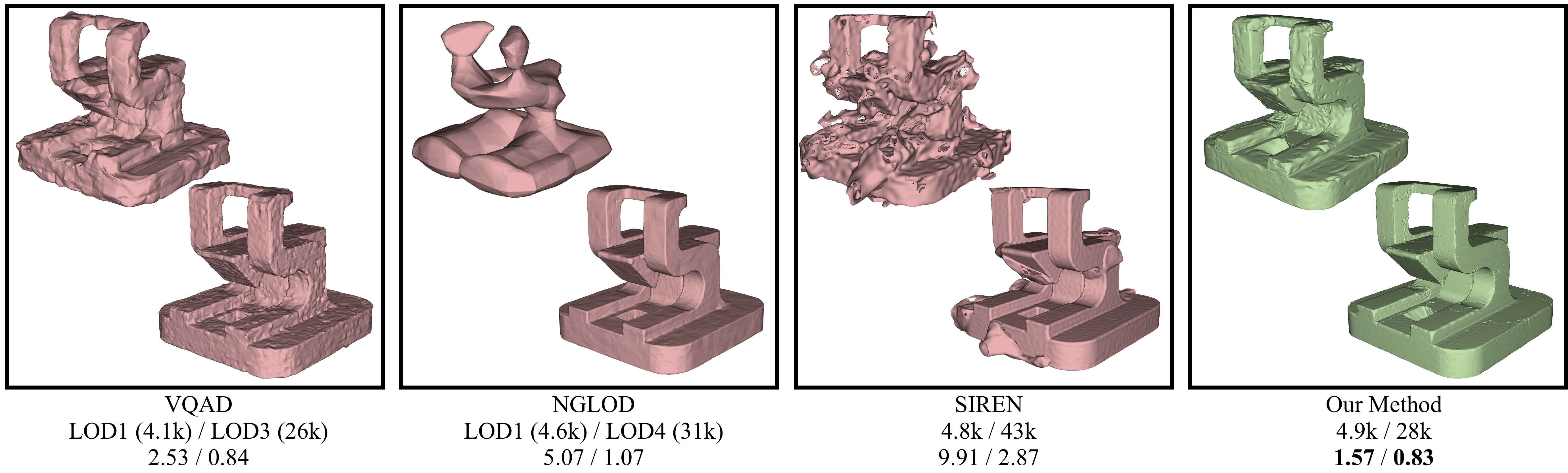}
    \includegraphics[width=\linewidth, trim=90 370 240 230, clip]{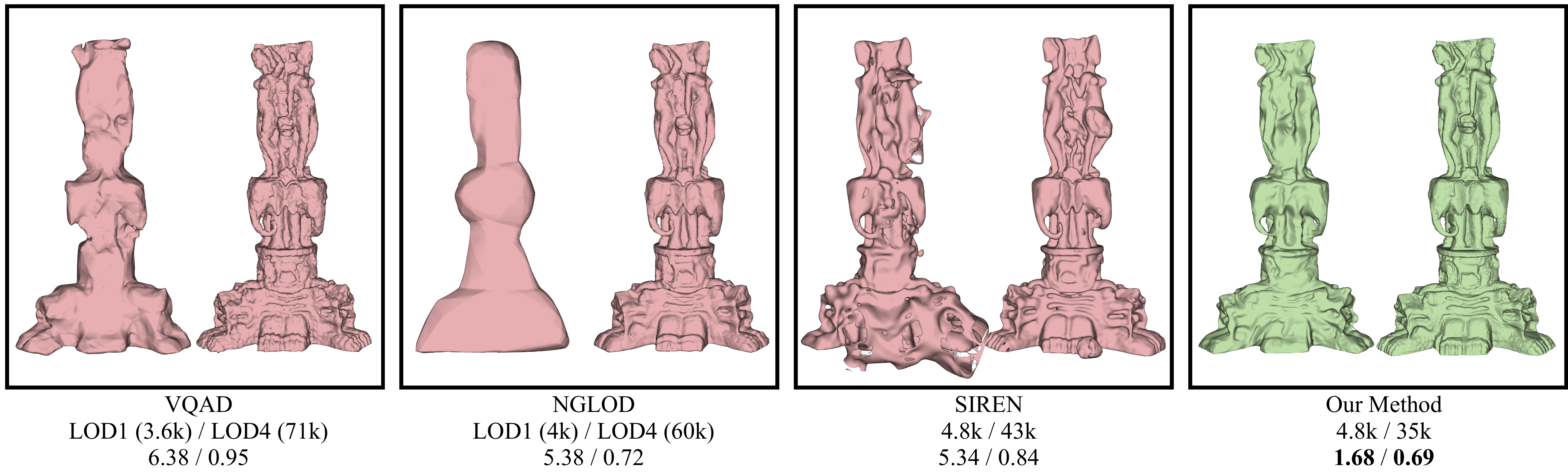}
    \vspace{-2em}
    \caption{{\bf Qualitative highlights:}
            Renderings for the `Lucy simplified' (top), `Shape 60' (middle), and `Thai statue' (bottom) inputs.
            We mark the parameter count for each method under each shape, as well as their Chamfer-$\Lone$ distances (multiplied by $10^3$).
            Our method provides the most detailed reconstruction compared to all alternatives, across all parameter counts.
            Notice the fine details at parameter counts as low as 4.8k.
    }
    \label{fig:qual}
\end{figure*}
\begin{figure*}
    \centering
    \begin{subfigure}[b]{0.245\linewidth}
        \includegraphics[width=\linewidth]{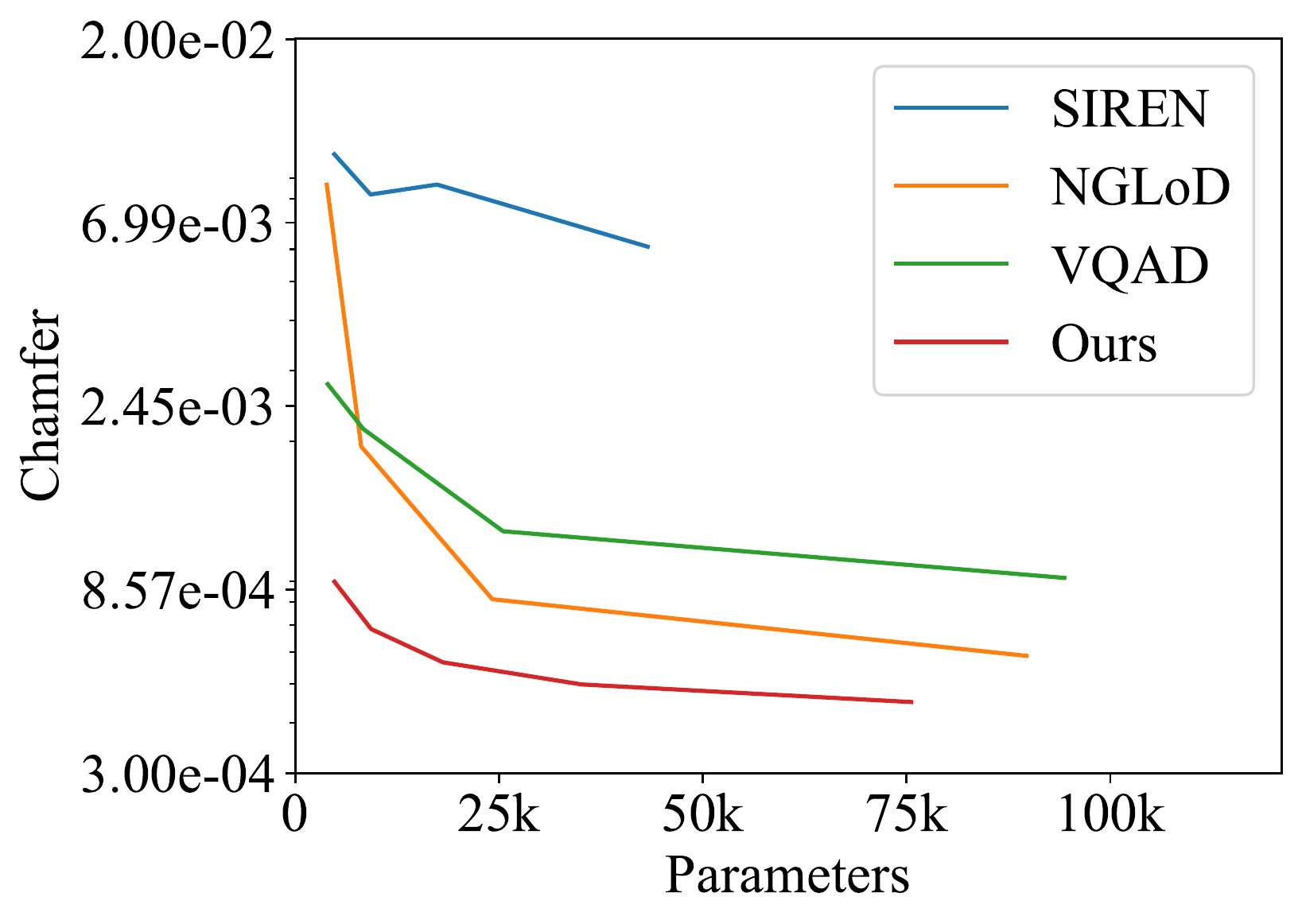}
        \vspace{-1.5em}
        \caption{Thingi10k}
    \end{subfigure}
    \begin{subfigure}[b]{0.245\linewidth}
        \includegraphics[width=\linewidth]{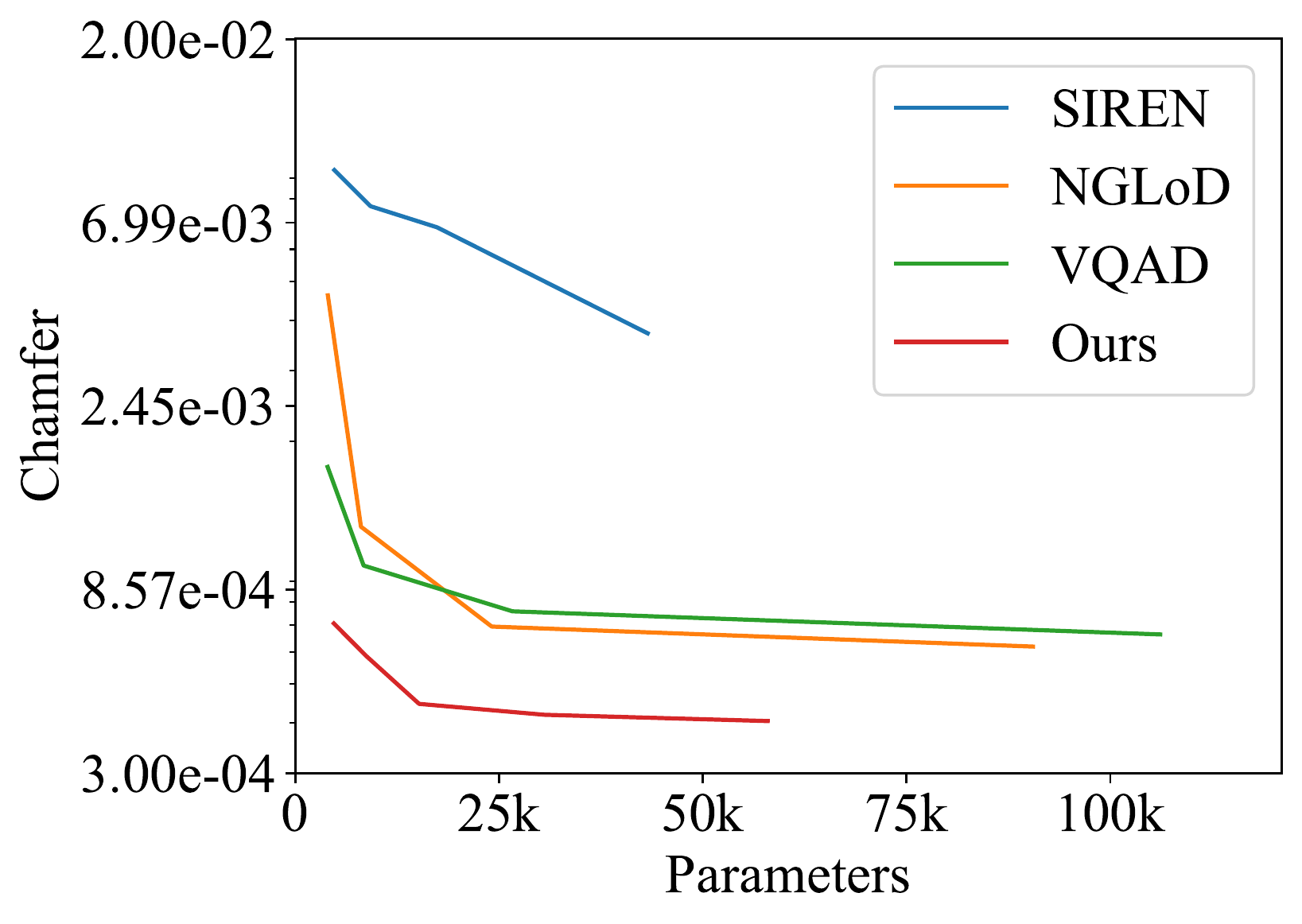}
        \vspace{-1.5em}
        \caption{ABC}
    \end{subfigure}
    \begin{subfigure}[b]{0.245\linewidth}
        \includegraphics[width=\linewidth]{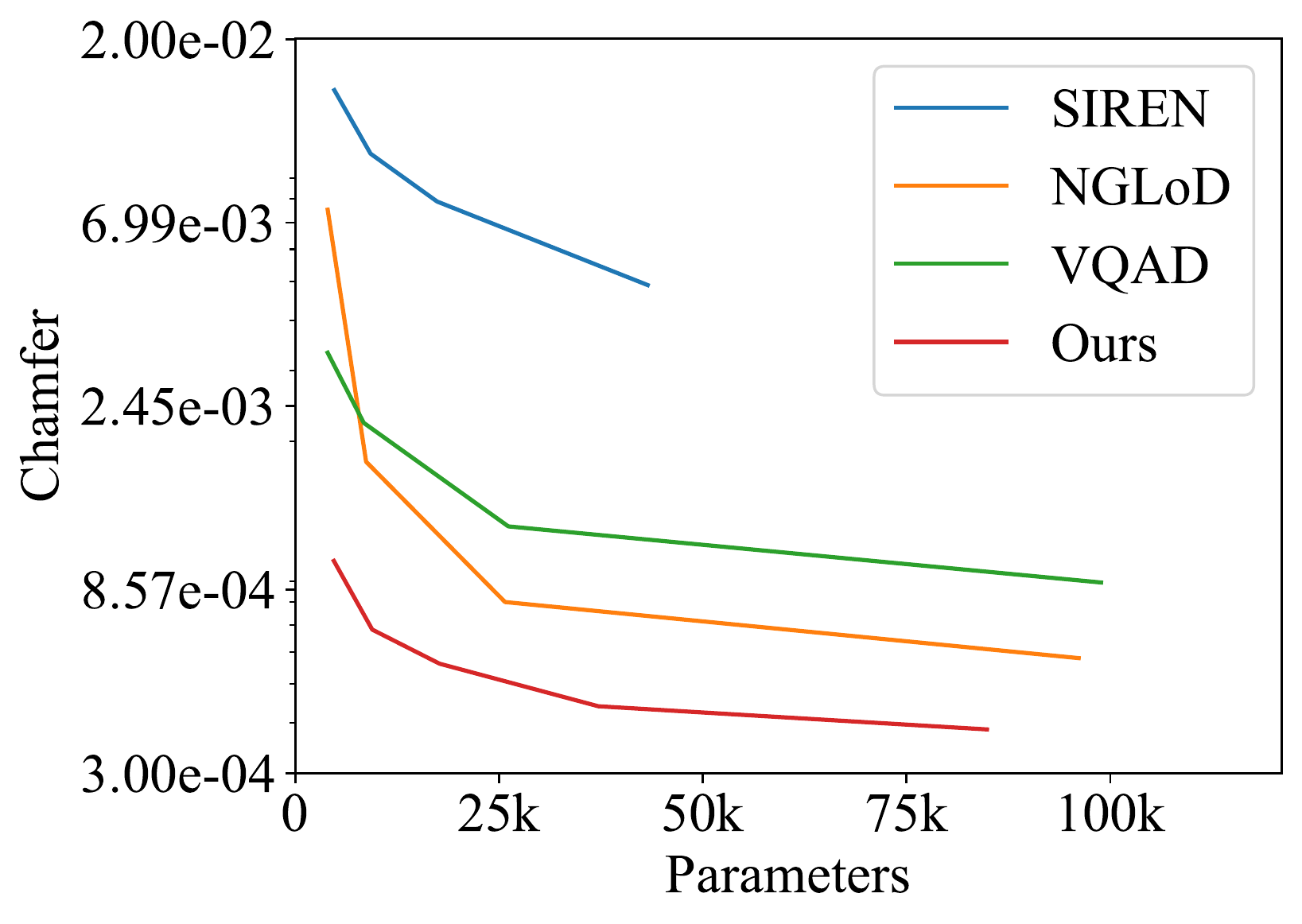}
        \vspace{-1.5em}
        \caption{DHFSlicer+Stanford}
    \end{subfigure}
    \begin{subfigure}[b]{0.245\linewidth}
        \includegraphics[width=\linewidth]{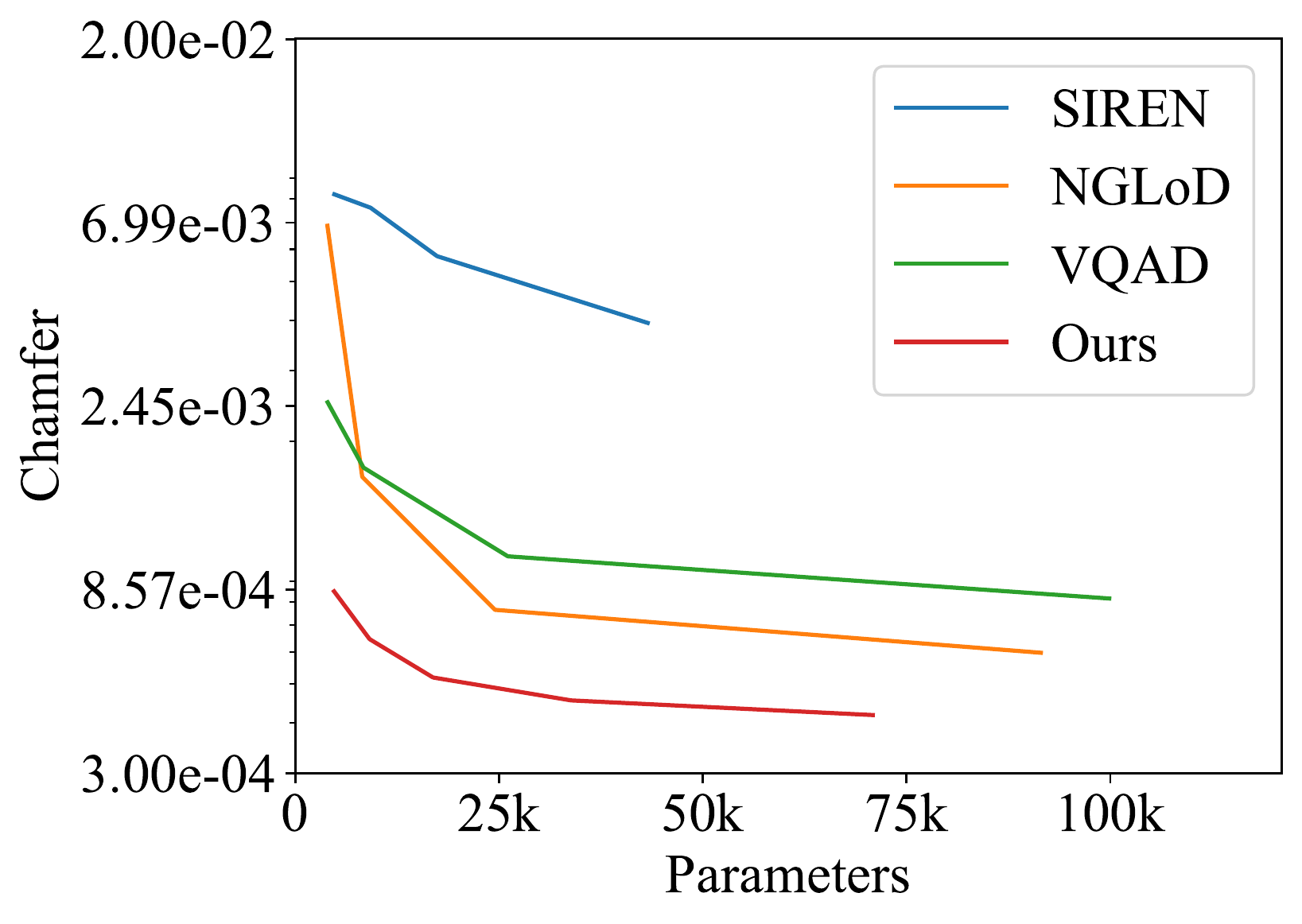}
        \vspace{-1.5em}
        \caption{Overall}
    \end{subfigure}
    \vspace{-1em}
    \caption{{\bf Quantitative summary:}
            Average Chamfer $\Lone$ distance vs average parameter count for each method for each dataset subset.
            Our method, CN-DHF, consistently outperforms all baselines.
    }%
    \label{fig:quant}
\end{figure*}

\vspace{-\customparskip}
\paragraph{Qualitative Comparison.}
\Fig{qual}
visually compare our learned surfaces against those generated by the main alternatives.
All renders were generated via raytracing, and colored using
a flat shading scheme.
As these representative examples show, CN-DHF generates input encodings that are both more accurate and more detailed compared to all alternatives and across different parameter counts.
We also noted that training SIREN with low parameter counts created training instability, resulting in floaters; whereas training in 2D with our method does not.

\begin{table}
	\begin{center}
		\resizebox{\linewidth}{!}{
			\begin{tabular}{r c c c c}
				\toprule
				                                & SIREN & NGLOD & VQAD & CN-DHF     \\
				\midrule
				LOD1 / Approx. 5K\hspace{0.5em} & 11.07 & 6.91  & 2.41 & {\bf 0.81} \\
				LOD2 / Approx. 10K              & 8.66  & 1.62  & 1.69 & {\bf 0.63} \\
				LOD3 / Approx. 20K              & 8.00  & 0.76  & 1.02 & {\bf 0.5}  \\
				LOD4 / Approx. 40K              & 5.01  & 0.6   & 0.81 & {\bf 0.45} \\
				\midrule
				Overall                         & 8.18  & 2.47  & 1.48 & {\bf 0.6}  \\
				\bottomrule
			\end{tabular}
		}
	\end{center}
	\vspace{-1.5em}
	\caption{{\bf Quantitative comparison:} Average Chamfer $\Lone$ distances  (multiplied by $10^3$) for SIREN \cite{siren}, NGLOD \cite{nglod}, VQAD \cite{vqad},  and CN-DHF at comparable parameter counts/LOD, and overall. Our method consistently outperforms all alternatives.}
	\label{tab:allres}
\end{table}
\vspace{-\customparskip}
\paragraph{Quantitative Comparison.}
We quantiatively evaluate our method through the standard protocol~\cite{nglod} of measuring Chamfer-$\Lone$ distance.
We sample {5M} points on each surface via ray stabbing, following \cite{nglod,idf}. %

We summarize our results in \Fig{quant} and \Tab{allres}. A difficulty in comparing  against NGLOD and VQAD is that their parameter counts change from one shape to another.
We enable fair comparisons by using two types of measurements.
 \Fig{quant} summarizes the results of each method independently by averaging the number of parameter count for each parameter configuration for each method---for example, for NGLOD, each point corresponds to different level-of-detail configurations---and plotting those against the corresponding average Chamfer distance.
As shown, our method significantly outperforms all baselines when comparing the error given the same parameter count.

In \Tab{allres} we provide more detailed statistics.
Since the number of DHFs  we use to represent each input varies, when comparing results against alternatives we compare against models with similar total parameter count (e.g. comparing a CN-DHF model that uses two DHF networks with 5K parameters each against SIREN at 10K parameters and NGLOD/VQAD at LOD level 2).
As the results show, our method outperforms all alternatives at all parameter counts, with the difference most pronounced at lower parameter counts.
Notably, we outperform the state of the art not just on average but also on the vast majority of individual inputs: our models are more accurate than corresponding SIREN ones on all inputs, more accurate than corresponding VQAD ones on 385 out of 400 inputs, and more accurate than corresponding NGLOD ones on 376 out of 400 inputs.
Predictably, improvement is most pronounced on inputs which can be accurately approximated with fewer DHFs: on inputs that require only one DHF (47\% of the corpus) our method is $3.4$ times more accurate than VQAD, the best performing alternative; on inputs that require 2 DHFs (44\% of the corpus) the improvement is by factor $2.4$, while on the 8.5\% of inputs that require three DHFs the improvement is by factor $1.2$.

\paragraph{Comparison against Wang~\etal~\cite{idf}.}
 \Tab{vswang} compares our results against the method of Wang~\etal~\cite{idf} on the three inputs they present that are common to our data set.
At the highest parameter count we evaluate on, our method reduces Chamfer distance by an average factor of $1.86$.
At the same time, we achieve better file size than their method: while their models require a file size of $4.8$ megabytes, our encoded results required a file size of only $146$ kilobytes.

\subsection{Ablation Studies}
\label{sec:exp_ablation}

\paragraph{DHF Count.}
Our outputs are defined via intersection of DHF surfaces, each represented using a neural network.
Thus the number of parameters necessary to encode a shape up to a desired accuracy is linearly dependent on the number of DHF surfaces necessary to accurately approximate the input.
We ablate the number of DHFs necessary to accurately approximate typical shapes, by measuring the percentage of the visible surface of a range of shapes that is covered by a given number of DHFs computed using our method (\Sec{view}).
\Tab{viewdir} reports the numbers for 1 through 4 directions, as well as the percentage of the surface visible from any of our 53 candidate directions.
We report the numbers for our dataset, as well as 500 random inputs from Thingi10K which include models with different connected component counts and open surfaces, and a commonly used dataset of closed meshes \mylesdata (114 meshes).
As the numbers show, while the coverage improves as one goes from 1 to 3 DHFs the impact of adding more DHFs is negligible, motivating our cutoff.
An example of a complicated shape being represented accurately with only 2 views is shown in \Fig{dhf}.

\begin{table}
    \begin{center}
        \resizebox{\linewidth}{!}{
            \begin{tabular}{r cc cc cc }
                \toprule
                                      & \multicolumn{2}{c}{dragon}  & \multicolumn{2}{c}{Thai Statue} & \multicolumn{2}{c}{Lion}                                                                        \\
                \cmidrule(lr){2-3} \cmidrule(lr){4-5} \cmidrule(lr){6-7}
                                      & Chamfer-$\Lone$$\downarrow$ & Size$\downarrow$                & Chamfer-$\Lone$$\downarrow$ & Size$\downarrow$ & Chamfer-$\Lone$$\downarrow$ & Size$\downarrow$ \\
                \midrule
                CN-DHF                & {\bf 0.63}                  & {\bf 146KB}                     & {\bf 0.69}                  & {\bf 146KB}      & {\bf 0.63}                  & {\bf 146KB}      \\
                Wang~\etal~\cite{idf} & 1.24                        & 4.8MB                           & 1.07                        & 4.8MB            & 1.31                        & 4.8MB            \\
                \bottomrule
            \end{tabular}
        }
    \end{center}
    \vspace{-1.5em}
    \caption{{\bf Comparison to Wang~\etal~\cite{idf}:}
            Chamfer-$\Lone$ distance and the size of the model for common shapes in our datasets and Wang~\etal~\cite{idf}.
            Ours provides better reconstruction with a significantly more compact ($33 times$ smaller) model.
    }
    \label{tab:vswang}
\end{table}
\begin{table}
    \begin{center}
        \resizebox{\linewidth}{!}{
            \begin{tabular}{r c c c c  c }
                \toprule
                Dataset    & 1 DHF & 2 DHFs & 3 DHFs & 4 DHFs & Vis. \% \\
                \midrule
                CN-DHF     & 95.9  & 99.0   & 99.6   & 99.7   & 99.7    \\
                Thingi10K  & 94.3  & 98.3   & 99.2   & 99.5   & 98.1    \\
                \mylesdata & 94.3  & 98.6   & 99.4   & 99.6   & 99.8    \\
                \bottomrule
            \end{tabular}
        }
    \end{center}
    \vspace{-1.5em}
    \caption{{\bf Coverage as function of DHF count:} Left to right percentage of visible surface covered by 1 to 4 DHFs, overall percentage of surface visible from outside. Rows correspond to our dataset, 500 random inputs from Thingi10K, and dataset from \mylesdata.}
    \label{tab:viewdir}
\end{table}

\paragraph{File Size.} The file size of a CN-DHF is determined by two factors: parameter count, and the number of DHF directions used. Per DHF axis direction, and without additional compression, our SIREN encoding takes 23kb for a 5k parameter DHF, 41kb for a 10k parameter DHF, 73kb for a 20k parameter DHF, and 173k for a 50k parameter DHF.

\paragraph{Impact of $\loss{lap}$.} We ablate the impact of our numerical Laplacian error term by encoding the {\em feline} model, with 2 axes and 3.2k parameters, with and without $\loss{lap}$. Without Laplacian error, chamfer distance increases from $1.131 \times 10^{-3}$ with $\loss{lap}$ to $2.432 \times 10^{-3}$ without.

\paragraph{More Ablations.}
Additional ablation results can be found in the \supp.

\subsection{Discussion and Limitations}

Our method is, by construction, designed for objects that can be well represented by implicit surfaces, and cannot encode portions of a model geometry which are invisible from outside the shape. Since such portions almost never need to be rendered, this limitation is rarely relevant for practical settings. As our ablations \Tab{viewdir} show, only a tiny percentage of the surface is typically occluded for most typical virtual objects used in graphical and other applications.

\begin{figure}
    \includegraphics[width=\linewidth]{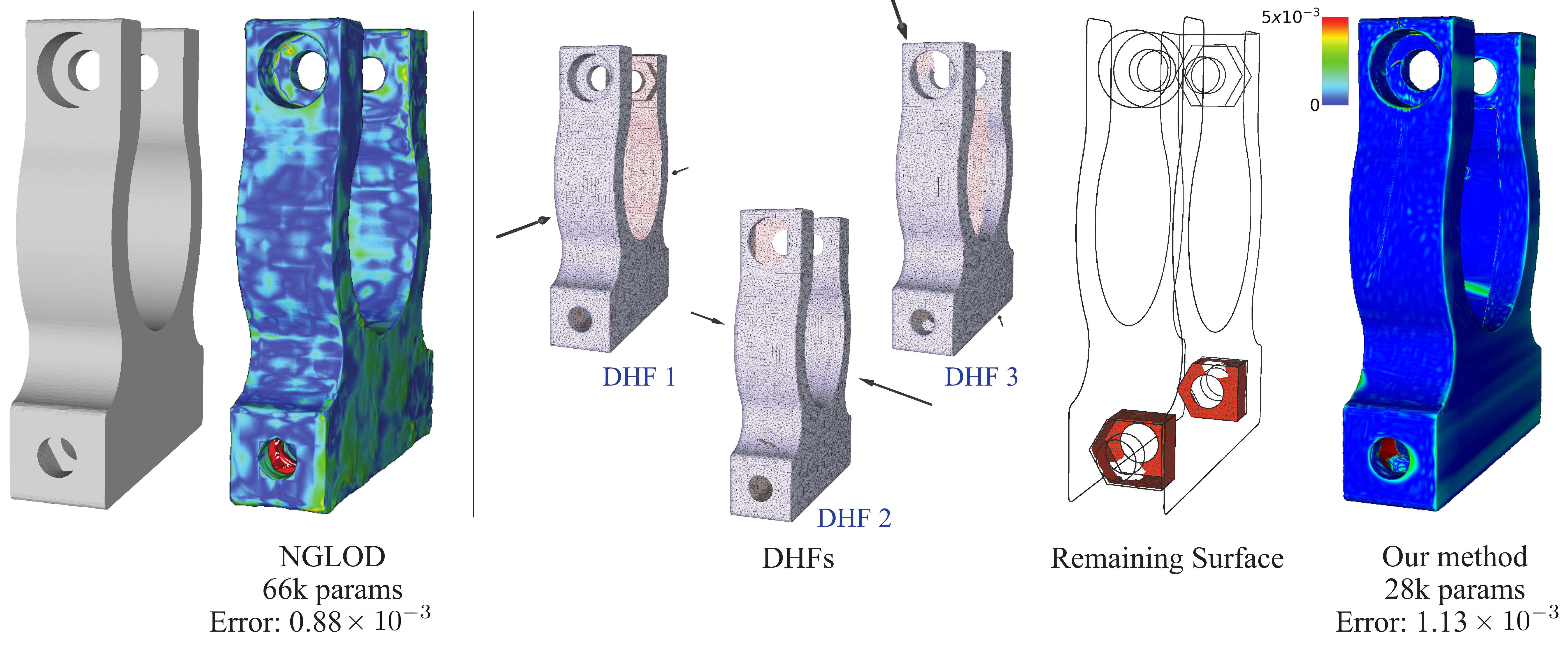}
    \caption{{\bf Limitations}: CN-DHF may fail to capture portions of input geometry that are poorly visible or invisible from outside the shape. On this example, we reproduce the outer surface of the shape more accurately than NGLOD, but NGLOD reproduces the inner geometry more accurately and consequently achieves improved overall accuracy.}
    \label{fig:limitations}
\end{figure}

As reported in \Tab{allres}, while rare, our method does not always provide the best reconstruction. In particular our reconstructions are inherently limited by the number of DHFs used, and on rare inputs where 3 DHFs are not sufficient to capture notable  features (\Fig{limitations}, other methods can outperform CN-DHF. %

\section{Conclusions}

We presented CN-DHF, a new hybrid neural implicit representation of 3D shapes, and demonstrated that it outperforms prior art in terms of compactness-to-accuracy tradeoff. Key to our representation is a novel use of DHF surfaces, and the formulation of shape approximation as an intersection of multiple DHFs. It would be interesting to explore if a variant of our method can be applied to shapes which contain occluded interior surfaces, as well as to use CN-DHFs for different generative tasks.

{\small
    \bibliographystyle{ieee_fullname}
    \bibliography{macros.bib, bibliography}

\begin{thebibliography}{10}\itemsep=-1pt

\bibitem{StanfordScanRep}
{The Stanford 3D Scanning Repository}.
\newblock \url{http://graphics.stanford.edu/data/3Dscanrep/}.

\bibitem{alderighi2021volume}
Thomas Alderighi, Luigi Malomo, Bernd Bickel, Paolo Cignoni, and Nico Pietroni.
\newblock Volume decomposition for two-piece rigid casting.
\newblock {\em ACM Transactions on Graphics}, 40(6):272--1, 2021.

\bibitem{Alemanno2014}
Giuseppe Alemanno, Paolo Cignoni, Nico Pietroni, Federico Ponchio, and Roberto
  Scopigno.
\newblock Interlocking pieces for printing tangible cultural heritage replicas.
\newblock In {\em Eurographics Workshop on Graphics and Cultural Heritage},
  pages 145--154, 2014.

\bibitem{BlinnImplicits}
James~F. Blinn.
\newblock A generalization of algebraic surface drawing.
\newblock {\em ACM Transactions on Graphics}, 1(3):235–256, jul 1982.

\bibitem{reverseengsurvey}
Francesco Buonamici, Monica Carfagni, Rocco Furferi, Lapo Governi, Alessandro
  Lapini, and Yary Volpe.
\newblock Reverse engineering modeling methods and tools: a survey.
\newblock {\em Computer-Aided Design and Applications}, 15(3):443--464, 2018.

\bibitem{Carr:2006:RMG}
Nathan~A. Carr, Jared Hoberock, Keenan Crane, and John~C. Hart.
\newblock Rectangular multi-chart geometry images.
\newblock In {\em Proceedings of the Eurographics symposium on Geometry
  processing}, 2006.

\bibitem{ChenSDFs}
Jianer Chen and Ergun Akleman.
\newblock Generalized distance functions.
\newblock In {\em Proceedings of the International Conference on Shape Modeling
  and Applications}, page~72, Los Alamitos, CA, USA, mar 1999. IEEE Computer
  Society.

\bibitem{bspnet}
Zhiqin Chen, Andrea Tagliasacchi, and Hao Zhang.
\newblock Bsp-net: Generating compact meshes via binary space partitioning.
\newblock {\em Proceedings of the IEEE/CVF Conference on Computer Vision and
  Pattern Recognition}, 2020.

\bibitem{imnet}
Zhiqin Chen and Hao Zhang.
\newblock Learning implicit fields for generative shape modeling.
\newblock In {\em Proceedings of the IEEE/CVF Conference on Computer Vision and
  Pattern Recognition}, pages 5939--5948, 2019.

\bibitem{davies2020overfit}
Thomas Davies, Derek Nowrouzezahrai, and Alec Jacobson.
\newblock Overfit neural networks as a compact shape representation.
\newblock {\em arXiv Preprint}, 2020.

\bibitem{deprelle2022learning}
Theo Deprelle, Thibault Groueix, Noam Aigerman, Vladimir~G Kim, and Mathieu
  Aubry.
\newblock Learning joint surface atlases.
\newblock {\em arXiv Preprint}, 2022.

\bibitem{deprelle2019learning}
Theo Deprelle, Thibault Groueix, Matthew Fisher, Vladimir Kim, Bryan Russell,
  and Mathieu Aubry.
\newblock Learning elementary structures for 3d shape generation and matching.
\newblock In {\em Advances in Neural Information Processing Systems}, pages
  7433--7443, 2019.

\bibitem{FeketeMiller}
Sándor Fekete and Joseph Mitchell.
\newblock Terrain decomposition and layered manufacturing.
\newblock {\em International Journal of Computational Geometry and
  Applications}, 11(06), 2001.

\bibitem{fu2008upright}
Hongbo Fu, Daniel Cohen-Or, Gideon Dror, and Alla Sheffer.
\newblock Upright orientation of man-made objects.
\newblock In {\em ACM SIGGRAPH}, pages 1--7. 2008.

\bibitem{gao2015revomaker}
Wei Gao, Yunbo Zhang, Diogo Nazzetta, Karthik Ramani, and Raymond Cipra.
\newblock Revomaker: Enabling multi-directional and functionally-embedded 3d
  printing using a rotational cuboidal platform.
\newblock In {\em User Insterface Software and Technology Symposium}. ACM,
  2015.

\bibitem{groueix2018}
Thibault Groueix, Matthew Fisher, Vladimir~G. Kim, Bryan Russell, and Mathieu
  Aubry.
\newblock {AtlasNet: A Papier-M\^ach\'e Approach to Learning 3D Surface
  Generation}.
\newblock In {\em Proceedings of the IEEE Conf. on Computer Vision and Pattern
  Recognition}, 2018.

\bibitem{geometryimages}
Xianfeng Gu, Steven~J Gortler, and Hugues Hoppe.
\newblock Geometry images.
\newblock In {\em Proceedings of the confernce on Computer graphics and
  interactive techniques}, pages 355--361, 2002.

\bibitem{MeshCNN}
Rana Hanocka, Amir Hertz, Noa Fish, Raja Giryes, Shachar Fleishman, and Daniel
  Cohen-Or.
\newblock Meshcnn: a network with an edge.
\newblock {\em ACM Transactions on Graphics}, 38(4):1--12, 2019.

\bibitem{herholz2015approximating}
Philipp Herholz, Wojciech Matusik, and Marc Alexa.
\newblock Approximating free-form geometry with height fields for
  manufacturing.
\newblock {\em Computer Graphics Forum}, 34(2), 2015.

\bibitem{Hu}
Ruizhen Hu, Honghua Li, Hao Zhang, and Daniel Cohen-Or.
\newblock Approximate pyramidal shape decomposition.
\newblock {\em ACM Transactions on Graphics}, 33(6), 2014.

\bibitem{jones20063d}
Mark~W Jones, J~Andreas Baerentzen, and Milos Sramek.
\newblock 3d distance fields: A survey of techniques and applications.
\newblock {\em IEEE Transactions on Visualization and Computer Graphics},
  12(4):581--599, 2006.

\bibitem{Keinert:2015}
Benjamin Keinert, Matthias Innmann, Michael S\"{a}nger, and Marc Stamminger.
\newblock Spherical fibonacci mapping.
\newblock {\em ACM Transactions on Graphics}, 34(6), 2015.

\bibitem{abcdataset}
Sebastian Koch, Albert Matveev, Zhongshi Jiang, Francis Williams, Alexey
  Artemov, Evgeny Burnaev, Marc Alexa, Denis Zorin, and Daniele Panozzo.
\newblock Abc: A big cad model dataset for geometric deep learning.
\newblock In {\em Proceedings of the IEEE/CVF Conference on Computer Vision and
  Pattern Recognition}, June 2019.

\bibitem{LH07}
Sylvain Lefebvre and Hugues Hoppe.
\newblock Compressed random-access trees for spatially coherent data.
\newblock In {\em Rendering Techniques (Proceedings of the Eurographics
  Symposium on Rendering)}. Eurographics, 2007.

\bibitem{li2023dense}
Heng Li, Xiaodong Gu, Weihao Yuan, Luwei Yang, Zilong Dong, and Ping Tan.
\newblock Dense rgb slam with neural implicit maps.
\newblock {\em International Conference on Learning Representations}, 2023.

\bibitem{maron}
Haggai Maron, Meirav Galun, Noam Aigerman, Miri Trope, Nadav Dym, Ersin Yumer,
  Vladimir~G Kim, and Yaron Lipman.
\newblock Convolutional neural networks on surfaces via seamless toric covers.
\newblock {\em ACM Transactions on Graphics}, 36(4):71--1, 2017.

\bibitem{occnet}
Lars Mescheder, Michael Oechsle, Michael Niemeyer, Sebastian Nowozin, and
  Andreas Geiger.
\newblock Occupancy networks: Learning 3d reconstruction in function space.
\newblock In {\em Proceedings of the IEEE/CVF Conference on Computer Vision and
  Pattern Recognition}, pages 4460--4470, 2019.

\bibitem{instantngp}
Thomas M{\"u}ller, Alex Evans, Christoph Schied, and Alexander Keller.
\newblock Instant neural graphics primitives with a multiresolution hash
  encoding.
\newblock {\em ACM Transactions on Graphics}, 41(4):1--15, 2022.

\bibitem{Muntoni2018}
Alessandro Muntoni, Marco Livesu, Riccardo Scateni, Alla Sheffer, and Daliele
  Panozzo.
\newblock Axis-aligned height-field block decomposition of 3d shapes.
\newblock {\em ACM Transactions on Graphics}, 37(5):169:1--169:15, 2018.

\bibitem{Muntoni2019}
Alessandro Muntoni, Lucio~Davide Spano, and Riccardo Scateni.
\newblock {Split and Mill: User Assisted Height-field Block Decomposition for
  Fabrication}.
\newblock In {\em Smart Tools and Apps for Graphics}, 2019.

\bibitem{musethvdb}
Ken Museth.
\newblock Vdb: High-resolution sparse volumes with dynamic topology.
\newblock {\em ACM Transactions on Graphics}, 32(3), jul 2013.

\bibitem{Myles16}
Ashish Myles, Nico Pietroni, and Denis Zorin.
\newblock Robust field-aligned global parametrization.
\newblock {\em ACM Transactions on Graphics}, 33(4):Article No. 135, 2014.

\bibitem{novak2012rasterized}
Jan Nov{\'a}k and Carsten Dachsbacher.
\newblock Rasterized bounding volume hierarchies.
\newblock In {\em Computer Graphics Forum}, volume~31, pages 403--412. Wiley
  Online Library, 2012.

\bibitem{unisurf}
Michael Oechsle, Songyou Peng, and Andreas Geiger.
\newblock Unisurf: Unifying neural implicit surfaces and radiance fields for
  multi-view reconstruction.
\newblock In {\em Proceedings of the IEEE/CVF Conference on Computer Vision and
  Pattern Recognition}, pages 5589--5599, 2021.

\bibitem{osher2004level}
Stanley Osher and Ronald~P Fedkiw.
\newblock Level set methods and dynamic implicit surfaces.
\newblock 2005.

\bibitem{deepsdf}
Jeong~Joon Park, Peter Florence, Julian Straub, Richard Newcombe, and Steven
  Lovegrove.
\newblock Deep{SDF}: Learning continuous signed distance functions for shape
  representation.
\newblock In {\em Proceedings of the IEEE/CVF Conference on Computer Vision and
  Pattern Recognition}, pages 165--174, 2019.

\bibitem{peng2005technologies}
Jingliang Peng, Chang-Su Kim, and C-C~Jay Kuo.
\newblock Technologies for 3d mesh compression: A survey.
\newblock {\em Journal of Visual Communication and Image Representation},
  16(6):688--733, 2005.

\bibitem{saito2019}
Shunsuke Saito, Zeng Huang, Ryota Natsume, Shigeo Morishima, Hao Li, and Angjoo
  Kanazawa.
\newblock Pifu: Pixel-aligned implicit function for high-resolution clothed
  human digitization.
\newblock In {\em International Conference on Computer Vision}, pages
  2304--2314, 2019.

\bibitem{sander2003multi}
Pedro~V Sander, Zo{\"e}~J Wood, Steven Gortler, John Snyder, and Hugues Hoppe.
\newblock Multi-chart geometry images.
\newblock 2003.

\bibitem{schmidt2005sketch}
Ryan Schmidt, Brian Wyvill, and Mario~Costa Sousa.
\newblock Sketch-based modeling with the blob tree.
\newblock In {\em ACM SIGGRAPH}, pages 90--es. 2005.

\bibitem{sheffer2007mesh}
Alla Sheffer, Emil Praun, Kenneth Rose, et~al.
\newblock Mesh parameterization methods and their applications.
\newblock {\em Foundations and Trends in Computer Graphics and Vision},
  2(2):105--171, 2007.

\bibitem{sitzmann2019metasdf}
Vincent Sitzmann, Eric~R. Chan, Richard Tucker, Noah Snavely, and Gordon
  Wetzstein.
\newblock Metasdf: Meta-learning signed distance functions.
\newblock In {\em arXiv Preprint}, 2020.

\bibitem{siren}
Vincent Sitzmann, Julien~N.P. Martel, Alexander~W. Bergman, David~B. Lindell,
  and Gordon Wetzstein.
\newblock Implicit neural representations with periodic activation functions.
\newblock In {\em Proceedings of the Advances in Neural Information Processing
  Systems}, 2020.

\bibitem{canonicalcapsules}
Weiwei Sun, Andrea Tagliasacchi, Boyang Deng, Sara Sabour, Soroosh Yazdani,
  Geoffrey~E Hinton, and Kwang~Moo Yi.
\newblock Canonical capsules: Self-supervised capsules in canonical pose.
\newblock {\em Advances in Neural Information Processing Systems},
  34:24993--25005, 2021.

\bibitem{vqad}
Towaki Takikawa, Alex Evans, Jonathan Tremblay, Thomas M{\"u}ller, Morgan
  McGuire, Alec Jacobson, and Sanja Fidler.
\newblock Variable bitrate neural fields.
\newblock In {\em ACM SIGGRAPH Conference Proceedings}, pages 1--9, 2022.

\bibitem{Takikawa2022SDF}
Towaki Takikawa, Andrew Glassner, and Morgan McGuire.
\newblock A dataset and explorer for 3d signed distance functions.
\newblock {\em Journal of Computer Graphics Techniques}, 11(2):1--29, April
  2022.

\bibitem{nglod}
Towaki Takikawa, Joey Litalien, Kangxue Yin, Karsten Kreis, Charles Loop, Derek
  Nowrouzezahrai, Alec Jacobson, Morgan McGuire, and Sanja Fidler.
\newblock Neural geometric level of detail: Real-time rendering with implicit
  3d shapes.
\newblock In {\em Proceedings of the IEEE/CVF Conference on Computer Vision and
  Pattern Recognition}, pages 11358--11367, 2021.

\bibitem{toumagotsman}
Costa Touma and Craig Gotsman.
\newblock Triangle mesh compression.
\newblock In {\em Proceedings of the Graphics Interface}, pages 26--34.
  Canadian Information Processing Society, 1998.

\bibitem{neus}
Peng Wang, Lingjie Liu, Yuan Liu, Christian Theobalt, Taku Komura, and Wenping
  Wang.
\newblock Neus: Learning neural implicit surfaces by volume rendering for
  multi-view reconstruction.
\newblock {\em Advances in Neural Information Processing Systems}, 2021.

\bibitem{wyvill1998blob}
Brian Wyvill, Andrew Guy, and Eric Galin.
\newblock The blob tree.
\newblock {\em Journal of Implicit Surfaces}, 3, 1998.

\bibitem{vincentsurvey}
Yiheng Xie, Towaki Takikawa, Shunsuke Saito, Or Litany, Shiqin Yan, Numair
  Khan, Federico Tombari, James Tompkin, Vincent Sitzmann, and Srinath Sridhar.
\newblock Neural fields in visual computing and beyond.
\newblock {\em Computer Graphics Forum}, 2022.

\bibitem{yang2020dhfslicer}
Jinfan Yang, Chrystiano Araujo, Nicholas Vining, Zachary Ferguson, Enrique
  Rosales, Daniele Panozzo, Sylvain Lefebvre, Paolo Cignoni, and Alla Sheffer.
\newblock Dhfslicer: Double height-field slicing for milling fixed-height
  materials.
\newblock {\em ACM Transactions on Graphics}, 2020.

\bibitem{volsdf}
Lior Yariv, Jiatao Gu, Yoni Kasten, and Yaron Lipman.
\newblock Volume rendering of neural implicit surfaces.
\newblock In {\em Advances in Neural Information Processing Systems}, 2021.

\bibitem{idf}
Wang Yifan, Lukas Rahmann, and Olga Sorkine-hornung.
\newblock Geometry-consistent neural shape representation with implicit
  displacement fields.
\newblock In {\em International Conference on Learning Representations}, 2022.

\bibitem{zhang2021}
Kai Zhang, Fujun Luan, Qianqian Wang, Kavita Bala, and Noah Snavely.
\newblock {PhySG}: {I}nverse rendering with spherical gaussians for
  physics-based material editing and relighting.
\newblock In {\em Conference on Computer Vision and Pattern Recognition}, 2021.

\bibitem{thingi10k}
Qingnan Zhou and Alec Jacobson.
\newblock Thingi10k: A dataset of 10,000 3d-printing models.
\newblock {\em arXiv Preprint}, 2016.

\bibitem{zhu2021rgb}
Luyang Zhu, Arsalan Mousavian, Yu Xiang, Hammad Mazhar, Jozef van Eenbergen,
  Shoubhik Debnath, and Dieter Fox.
\newblock Rgb-d local implicit function for depth completion of transparent
  objects.
\newblock In {\em Proceedings of the IEEE/CVF Conference on Computer Vision and
  Pattern Recognition}, pages 4649--4658, 2021.

\end{thebibliography}
}

\clearpage

\appendix
\renewcommand\thefigure{\Alph{figure}}
\setcounter{figure}{0}
\renewcommand\thetable{\Alph{table}}
\setcounter{table}{0}
\setcounter{footnote}{0}
\twocolumn[
    \centering
    \Large
    \textbf{CN-DHF: Compact Neural Double Height-Field Representations of 3D Shapes} \\
    \vspace{0.5em} (Supplementary Material) \\
    \vspace{1.0em}
]

\section{More ablation results}

To validate our design choices we perform ablation studies using a total of five representative shapes. We use `0032' and `9942' from the ABC dataset, `Lucy Simplified' and `Bunny' from Thingi10k, and `Happy Buddha' from the DHFSlicer and Stanford datasets.

\paragraph{Height Field Value for the Outside.}
In \eq{lossreg} we simply enforce that the order of the two estimated height field values is correct for locations that correspond to the outside of the shape.
In addition to this loss, we train an alternative strategy where we directly regress a conventional setup---having height field values set to be $-1$ or $+1$ for the outside.
As we report in \Tab{ablationloss}, regressing a fixed value results in significantly worse performance.
We attribute this to the fact that the network needs to learn to regress these values, which requires spending parts of the network capacity to do so.
This is unnecessary for our ordering-based loss in \eq{lossreg}.

\begin{figure}
    \centering
    \includegraphics[width=0.8\linewidth]{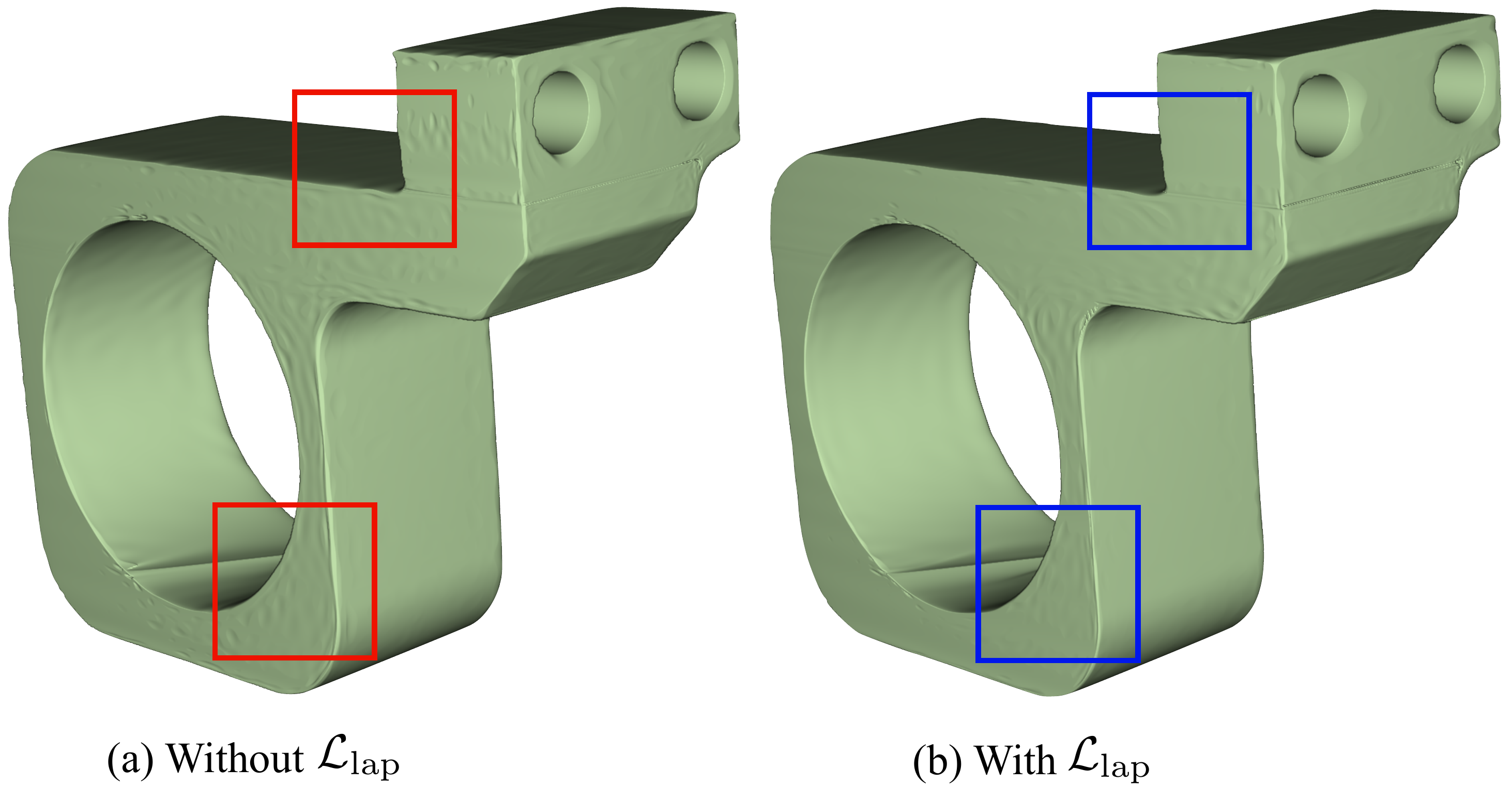}
    \vspace{-1em}
    \caption{%
        {\bf Ablation study -- smoothness loss $\loss{lap}$:}
        renderings of our model (a) with and (b) without the smoothness loss for the 9942 shape from the ABC dataset, with 9,602 parameters.
        Smoothness loss encourages the estimated surfaces to have \emph{correct} normals, leading to smoother surfaces as shown.
    }
    \label{fig:ablationlap}
\end{figure}

\paragraph{Smoothness loss -- $\loss{lap}$.}
Our loss function leverages the Laplacian encoding of the top and bottom DHF surfaces to encourage smoothness and proper normals.
We report Chamfer distances with and without the smoothness loss in \Tab{ablationloss}.
As reported, the loss term helps slightly when dealing with vey compact models, and provides similar results when enough capacity is provided.
The effect of this loss is better appreciated qualitatively in \Fig{ablationlap}, where we show the bumpiness of the surface being smoothed out.

\paragraph{Number of Views.}
We further study how the number of views affect reconstruction performance by quantitatively evaluating the reconstruction performance when we use one less or one more view than what our method in \Sec{view} provides. To obtain the extra view, we increase the coverage threshold to 100\% and add an additional view found by our greedy strategy. We report results in \Tab{numview}.

\section{All Results}
We provide detailed results for all shapes used for computing \Tab{allres} in \Tab{fullres}.
As mentioned earlier in \Sec{exp_quant}, each method returns different parameter counts for each shape, making the comparison difficult.
We thus compare against similar parameter counts.
When we do not have an equivalent parameter count, for example when much higher parameter count is available for the baselines than our maximum number of parameters, we simply compare with our highest parameter count---this actually is unfavorable to our method, but we found that thanks to the compactness of our method, in many cases we outperform the compared baselines despite this disadvantage. We also provide rendering of each learned model in the accompanied \texttt{7572-renderings.pdf}.

\begin{table}
    \begin{center}
        \setlength\tabcolsep{4pt}
        \resizebox{\linewidth}{!}{
            \begin{tabular}{l l r c c c}
                \toprule
                Dataset   & Shape Name                 & Num. Params. & Regressing $-1$/$+1$ & Without $\loss{lap}$ & Our method (full) \\
                \midrule
                ABC       & {0032}                     & 9602         & 0.693                & 0.294                & {\bf 0.275}       \\
                ABC       & {9942}                     & 9602         & 0.525                & 0.407                & {\bf 0.391}       \\
                Thingi10k & {Lucy Simplified (252119)} & 14403        & 1.18                 & {\bf 0.761}          & 0.768             \\
                Thingi10k & {Bunny (441708)}           & 9602         & 0.507                & 0.440                & {\bf 0.428}       \\
                DHFSlicer & {Happy Buddha (happy)}     & 14403        & 1.12                 & 0.868                & {\bf 0.866}       \\
                \midrule[0.1pt]
                Overall   &                            & 11522        & 0.803                & 0.554                & {\bf 0.546}       \\
                \midrule
                ABC       & {0032}                     & 86482        & 0.254                & 0.261                & {\bf 0.231}       \\
                ABC       & {9942}                     & 86482        & 0.362                & {\bf 0.324}          & 0.325             \\
                Thingi10k & {Lucy Simplified (252119)} & 129723       & 0.389                & 0.357                & {\bf 0.350}       \\
                Thingi10k & {Bunny (441708)}           & 86482        & 0.304                & 0.{\bf 302}          & 0.303             \\
                DHFSlicer & {Happy Buddha (happy)}     & 129723       & 0.503                & 0.{\bf 462}          & 0.476             \\
                \midrule[0.1pt]
                Overall   &                            & 103778       & 0.362                & 0.341                & {\bf 0.337}       \\
                \bottomrule
            \end{tabular}
        }
    \end{center}
    \vspace{-1em}
    \caption{
        {\bf Ablation -- loss function:}
        We report the Chamfer-$\Lone$ distance multiplied by $10^3$ with our full model, without the smoothness loss, and when regressing $-1$/$+1$ for outside.
        Smoothness loss helps most in simple shapes and with low parameter counts.
        Regressing specific values significantly deteriorates performance.
        Our formalation performs best.
    }
    \label{tab:ablationloss}
\end{table}
\begin{table}
    \begin{center}
        \setlength\tabcolsep{4pt}
        \resizebox{\linewidth}{!}{
            \begin{tabular}{l l c c c}
                \toprule
                Dataset   & Shape Name                 & One less view & Our method  & One more view \\
                \midrule
                ABC       & {0032}                     & 0.982         & {\bf 0.231} & ---           \\
                ABC       & {9942}                     & 0.840         & {\bf 0.325} & ---           \\
                Thingi10k & {Lucy Simplified (252119)} & 0.397         & {\bf 0.350} & 0.367         \\
                Thingi10k & {Bunny (441708)}           & 0.383         & {\bf 0.303} & 0.310         \\
                DHFSlicer & {Happy Buddha (happy)}     & 0.630         & 0.476       & {\bf 0.466}   \\
                \bottomrule
            \end{tabular}
        }
    \end{center}
    \vspace{-1em}
    \caption{
        {\bf Ablation -- number of views:}
        We report the Chamfer-$\Lone$ distance multiplied by $10^3$ with varying number of views.
        Having fewer views significantly worsens the reconstructino performance.
        In case of the two shapes from the ABC dataset, the number of views from our method provides full surface coverage, hence additional views are unavailable.
        For the other three shapes, adding more views does not cause significant changes as the---it, in fact, sometimes slightly deteriorates performance due to minor errors in additional views interfering.
    }
    \label{tab:numview}
\end{table}

\onecolumn
{
    \footnotesize
    \setlength\tabcolsep{2pt}
    \begin{longtable}{lr|crc|crc|rc|rc}
        \caption{%
            {\bf Detailed quantitative results:}
            We report detailed results for all shapes and all configurations summarized in \Tab{allres}.
        }%
        \label{tab:fullres}
        \\
        \toprule                                                                              &      & \multicolumn{3}{c}{NGLOD} & \multicolumn{3}{c}{VQAD} & \multicolumn{2}{c}{SIREN} & \multicolumn{2}{c}{CN-DHF}                                                                            \\
        \cmidrule(lr){3-5} \cmidrule(lr){6-8} \cmidrule(lr){9-10} \cmidrule(lr){11-12}dataset & name & LOD                       & params                   & Chamfer $L_1$             & LOD                        & params & Chamfer $L_1$ & params & Chamfer $L_1$ & params & Chamfer $L_1$ \\
        \midrule \endfirsthead
        \toprule                                                                              &      & \multicolumn{3}{c}{NGLOD} & \multicolumn{3}{c}{VQAD} & \multicolumn{2}{c}{SIREN} & \multicolumn{2}{c}{CN-DHF}                                                                            \\
        \cmidrule(lr){3-5} \cmidrule(lr){6-8} \cmidrule(lr){9-10} \cmidrule(lr){11-12}dataset & name & LOD                       & params                   & Chamfer $L_1$             & LOD                        & params & Chamfer $L_1$ & params & Chamfer $L_1$ & params & Chamfer $L_1$ \\
        \midrule \endhead
        \bottomrule \endlastfoot
        \bottomrule \footnotesize{\emph{(continued\ldots)}} \endfoot
        ABC & 00000002 & 1 & 3905 & 3.0 & 1 & 4725 & 1.56 & 4801 & 7.91 & 4851 & \textbf{1.18} \\
ABC & 00000002 & 2 & 9953 & 0.97 & 2 & 12795 & 1.07 & 9241 & 5.21 & 9602 & \textbf{0.9} \\
ABC & 00000002 & 3 & 38129 & \textbf{0.51} & 3 & 44891 & 0.75 & 17401 & 5.24 & 14403 & 0.63\\
ABC & 00000002 & 4 & 162609 & \textbf{0.53} & 4 & 178377 & 0.61 & 43241 & 5.1 & 27723 & 0.55\\
ABC & 00000003 & 1 & 4577 & 2.33 & 1 & 4656 & 1.56 & 4801 & 3.09 & 4801 & \textbf{0.37} \\
ABC & 00000003 & 2 & 11553 & 0.66 & 2 & 11984 & 0.95 & 9241 & 1.85 & 9241 & \textbf{0.33} \\
ABC & 00000003 & 3 & 37617 & 0.51 & 3 & 46301 & 0.75 & 17401 & 4.89 & 17401 & \textbf{0.32} \\
ABC & 00000003 & 4 & 148801 & 0.55 & 4 & 198675 & 0.72 & 43241 & 0.87 & 17401 & \textbf{0.32} \\
ABC & 00000004 & 1 & 3809 & 3.44 & 1 & 3410 & 1.55 & 4801 & 1.14 & 4801 & \textbf{0.19} \\
ABC & 00000004 & 2 & 6321 & 1.12 & 2 & 4932 & 0.98 & 9241 & 0.48 & 9241 & \textbf{0.18} \\
ABC & 00000004 & 3 & 13217 & 0.5 & 3 & 9920 & 0.58 & 17401 & 0.71 & 17401 & \textbf{0.18} \\
ABC & 00000004 & 4 & 33473 & 0.42 & 4 & 32214 & 0.55 & 43241 & 0.59 & 17401 & \textbf{0.18} \\
ABC & 00000006 & 1 & 3905 & 1.78 & 1 & 4372 & 1.24 & 4801 & 8.93 & 4801 & \textbf{0.29} \\
ABC & 00000006 & 2 & 10065 & 0.66 & 2 & 10200 & 0.75 & 9241 & 9.55 & 4801 & \textbf{0.29} \\
ABC & 00000006 & 3 & 32513 & 0.54 & 3 & 35426 & 0.55 & 17401 & 4.88 & 4801 & \textbf{0.29} \\
ABC & 00000006 & 4 & 121425 & 0.57 & 4 & 135614 & 0.55 & 43241 & 6.75 & 43241 & \textbf{0.28} \\
ABC & 00000008 & 1 & 3905 & 3.74 & 1 & 4086 & 2.45 & 4801 & 0.83 & 4801 & \textbf{0.48} \\
ABC & 00000008 & 2 & 9857 & 1.54 & 2 & 9441 & 1.2 & 9241 & 0.65 & 9241 & \textbf{0.33} \\
ABC & 00000008 & 3 & 25809 & 1.07 & 3 & 32738 & 0.78 & 17401 & 3.12 & 17401 & \textbf{0.31} \\
ABC & 00000008 & 4 & 107377 & 0.52 & 4 & 144318 & 0.69 & 43241 & 0.51 & 43241 & \textbf{0.3} \\
ABC & 00000027 & 1 & 3857 & 6.71 & 1 & 3825 & 2.64 & 4801 & 33.4 & 3234 & \textbf{0.91} \\
ABC & 00000027 & 2 & 7665 & 1.66 & 2 & 7985 & 1.46 & 9241 & 2.46 & 9602 & \textbf{0.58} \\
ABC & 00000027 & 3 & 21201 & 0.76 & 3 & 25919 & 1.0 & 17401 & 12.79 & 18482 & \textbf{0.57} \\
ABC & 00000027 & 4 & 83585 & 0.67 & 4 & 101373 & 0.83 & 43241 & 16.55 & 34802 & \textbf{0.56} \\
ABC & 00000031 & 1 & 4705 & 2.27 & 1 & 4870 & 1.81 & 4801 & 16.74 & 4834 & \textbf{0.47} \\
ABC & 00000031 & 2 & 11697 & 1.02 & 2 & 13198 & 1.13 & 9241 & 16.5 & 9602 & \textbf{0.42} \\
ABC & 00000031 & 3 & 41921 & 0.65 & 3 & 50759 & 0.74 & 17401 & 12.93 & 18482 & \textbf{0.38} \\
ABC & 00000031 & 4 & 170321 & 0.64 & 4 & 193582 & 0.76 & 43241 & 5.2 & 34802 & \textbf{0.38} \\
ABC & 00000032 & 1 & 3425 & 3.33 & 1 & 3431 & 1.34 & 4801 & 1.1 & 4834 & \textbf{0.33} \\
ABC & 00000032 & 2 & 5441 & 1.33 & 2 & 5439 & 0.76 & 9241 & 0.63 & 9602 & \textbf{0.28} \\
ABC & 00000032 & 3 & 12897 & 0.93 & 3 & 13708 & 0.57 & 17401 & 0.57 & 18482 & \textbf{0.25} \\
ABC & 00000032 & 4 & 40929 & 0.59 & 4 & 47637 & 0.64 & 43241 & 2.87 & 34802 & \textbf{0.23} \\
ABC & 00000049 & 1 & 3425 & 4.54 & 1 & 3451 & 1.9 & 4801 & 1.47 & 4801 & \textbf{0.36} \\
ABC & 00000049 & 2 & 5313 & 2.2 & 2 & 5591 & 0.98 & 9241 & 4.47 & 9241 & \textbf{0.31} \\
ABC & 00000049 & 3 & 13409 & 1.36 & 3 & 14376 & 0.65 & 17401 & 0.65 & 17401 & \textbf{0.3} \\
ABC & 00000049 & 4 & 43585 & 0.6 & 4 & 51697 & 0.61 & 43241 & 2.99 & 17401 & \textbf{0.3} \\
ABC & 00000058 & 1 & 3809 & 4.28 & 1 & 3634 & 1.17 & 4801 & 5.98 & 4801 & \textbf{0.3} \\
ABC & 00000058 & 2 & 6529 & 0.96 & 2 & 6699 & 0.75 & 9241 & 5.43 & 9241 & \textbf{0.27} \\
ABC & 00000058 & 3 & 16913 & 0.48 & 3 & 17537 & 0.65 & 17401 & 2.62 & 17401 & \textbf{0.26} \\
ABC & 00000058 & 4 & 61777 & 0.57 & 4 & 62090 & 0.61 & 43241 & 0.63 & 17401 & \textbf{0.26} \\
ABC & 00000060 & 1 & 4561 & 5.07 & 1 & 4071 & 2.53 & 4801 & 9.91 & 4851 & \textbf{1.57} \\
ABC & 00000060 & 2 & 8193 & 1.63 & 2 & 9641 & 1.49 & 9241 & 19.64 & 9602 & \textbf{1.22} \\
ABC & 00000060 & 3 & 26097 & \textbf{0.85} & 3 & 30791 & 1.07 & 17401 & 7.39 & 14403 & 0.98\\
ABC & 00000060 & 4 & 111121 & \textbf{0.63} & 4 & 118394 & 0.68 & 43241 & 2.87 & 27723 & 0.83\\
ABC & 00000065 & 1 & 3905 & 4.48 & 1 & 3751 & 2.35 & 4801 & 4.4 & 3234 & \textbf{1.42} \\
ABC & 00000065 & 2 & 6945 & 1.86 & 2 & 6925 & 1.45 & 9241 & 4.33 & 9602 & \textbf{1.24} \\
ABC & 00000065 & 3 & 19841 & \textbf{0.94} & 3 & 21750 & 1.03 & 17401 & 4.08 & 18482 & 1.2\\
ABC & 00000065 & 4 & 66273 & 0.88 & 4 & 76465 & \textbf{0.76} & 43241 & 1.85 & 27723 & 1.13\\
ABC & 00000070 & 1 & 3713 & 9.65 & 1 & 3277 & 1.35 & 4801 & 0.52 & 4801 & \textbf{0.36} \\
ABC & 00000070 & 2 & 5505 & 0.98 & 2 & 4709 & 0.78 & 9241 & 2.27 & 9241 & \textbf{0.3} \\
ABC & 00000070 & 3 & 10801 & 0.49 & 3 & 10312 & 0.63 & 17401 & 0.56 & 17401 & \textbf{0.25} \\
ABC & 00000070 & 4 & 29633 & 0.54 & 4 & 33779 & 0.53 & 43241 & 0.53 & 17401 & \textbf{0.25} \\
ABC & 00000073 & 1 & 3889 & 6.29 & 1 & 3882 & 1.38 & 4801 & 21.29 & 4834 & \textbf{0.65} \\
ABC & 00000073 & 2 & 8641 & 1.12 & 2 & 8442 & 0.71 & 9241 & 1.99 & 9602 & \textbf{0.64} \\
ABC & 00000073 & 3 & 21729 & 0.81 & 3 & 21461 & 0.72 & 17401 & 3.56 & 9602 & \textbf{0.64} \\
ABC & 00000073 & 4 & 90529 & 0.62 & 4 & 84981 & 0.67 & 43241 & 4.51 & 34802 & \textbf{0.56} \\
ABC & 00000078 & 1 & 3425 & 4.14 & 1 & 3424 & 1.33 & 4801 & 0.68 & 4801 & \textbf{0.58} \\
ABC & 00000078 & 2 & 5937 & 0.91 & 2 & 5223 & 0.68 & 9241 & 6.31 & 9241 & \textbf{0.46} \\
ABC & 00000078 & 3 & 13425 & 0.48 & 3 & 11277 & 0.55 & 17401 & 1.68 & 17401 & \textbf{0.36} \\
ABC & 00000078 & 4 & 37377 & 0.46 & 4 & 34916 & 0.61 & 43241 & 1.59 & 43241 & \textbf{0.34} \\
ABC & 00000105 & 1 & 3425 & 2.87 & 1 & 3718 & 0.85 & 4801 & 4.4 & 4801 & \textbf{0.36} \\
ABC & 00000105 & 2 & 6721 & 0.63 & 2 & 7098 & 1.05 & 9241 & 5.02 & 9241 & \textbf{0.3} \\
ABC & 00000105 & 3 & 18993 & 0.57 & 3 & 19055 & 0.72 & 17401 & 3.67 & 17401 & \textbf{0.28} \\
ABC & 00000105 & 4 & 68833 & 0.68 & 4 & 76053 & 0.63 & 43241 & 2.06 & 17401 & \textbf{0.28} \\
ABC & 00000112 & 1 & 4705 & 5.37 & 1 & 5592 & 2.09 & 4801 & 11.51 & 4834 & \textbf{0.47} \\
ABC & 00000112 & 2 & 14673 & 0.88 & 2 & 17385 & 1.09 & 9241 & 10.87 & 9602 & \textbf{0.39} \\
ABC & 00000112 & 3 & 56065 & 0.54 & 3 & 68230 & 0.81 & 17401 & 11.26 & 18482 & \textbf{0.39} \\
ABC & 00000112 & 4 & 240513 & 0.56 & 4 & 266002 & 0.69 & 43241 & 7.27 & 34802 & \textbf{0.36} \\
ABC & 00000117 & 1 & 4193 & 3.95 & 1 & 4041 & 2.91 & 4801 & 7.9 & 4834 & \textbf{0.47} \\
ABC & 00000117 & 2 & 8897 & 2.46 & 2 & 9825 & 0.93 & 9241 & 6.81 & 9602 & \textbf{0.34} \\
ABC & 00000117 & 3 & 25537 & 0.63 & 3 & 32305 & 0.77 & 17401 & 6.02 & 18482 & \textbf{0.31} \\
ABC & 00000117 & 4 & 114049 & 0.5 & 4 & 134259 & 0.68 & 43241 & 3.15 & 34802 & \textbf{0.29} \\
ABC & 00000120 & 1 & 4705 & 3.83 & 1 & 4481 & 2.5 & 4801 & 28.16 & 4834 & \textbf{1.02} \\
ABC & 00000120 & 2 & 11633 & 0.74 & 2 & 9242 & 1.12 & 9241 & 9.22 & 9602 & \textbf{0.56} \\
ABC & 00000120 & 3 & 35313 & 0.5 & 3 & 26618 & 0.89 & 17401 & 2.51 & 18482 & \textbf{0.49} \\
ABC & 00000120 & 4 & 107393 & 0.5 & 4 & 99493 & 0.69 & 43241 & 1.2 & 34802 & \textbf{0.46} \\
ABC & 00000127 & 1 & 4561 & 4.83 & 1 & 4051 & 2.35 & 4801 & 13.57 & 4851 & \textbf{1.12} \\
ABC & 00000127 & 2 & 7393 & 1.78 & 2 & 9495 & 1.39 & 9241 & 4.6 & 4851 & \textbf{1.12} \\
ABC & 00000127 & 3 & 25105 & \textbf{0.8} & 3 & 31369 & 0.98 & 17401 & 9.48 & 14403 & 0.86\\
ABC & 00000127 & 4 & 108049 & \textbf{0.72} & 4 & 128314 & 0.82 & 43241 & 8.54 & 27723 & 0.76\\
ABC & 00000130 & 1 & 3905 & 9.05 & 1 & 3931 & 2.04 & 4801 & 39.16 & 4834 & \textbf{0.36} \\
ABC & 00000130 & 2 & 7041 & 1.36 & 2 & 8101 & 1.24 & 9241 & 41.35 & 9602 & \textbf{0.33} \\
ABC & 00000130 & 3 & 22817 & 0.62 & 3 & 26852 & 0.92 & 17401 & 0.53 & 18482 & \textbf{0.32} \\
ABC & 00000130 & 4 & 85585 & 0.61 & 4 & 111809 & 0.79 & 43241 & 1.23 & 34802 & \textbf{0.32} \\
ABC & 00000136 & 1 & 3425 & 2.86 & 1 & 3691 & 0.72 & 4801 & 1.34 & 4801 & \textbf{0.34} \\
ABC & 00000136 & 2 & 7041 & 0.94 & 2 & 6987 & 0.63 & 9241 & 6.79 & 9602 & \textbf{0.29} \\
ABC & 00000136 & 3 & 18401 & 0.75 & 3 & 19229 & 0.53 & 17401 & 33.1 & 18482 & \textbf{0.28} \\
ABC & 00000136 & 4 & 66913 & 0.57 & 4 & 67126 & 0.53 & 43241 & 6.04 & 34802 & \textbf{0.27} \\
ABC & 00000152 & 1 & 3905 & 12.78 & 1 & 4003 & 1.57 & 4801 & 6.02 & 4801 & \textbf{0.62} \\
ABC & 00000152 & 2 & 7041 & 1.21 & 2 & 9824 & 1.05 & 9241 & 7.92 & 9241 & \textbf{0.56} \\
ABC & 00000152 & 3 & 23969 & 0.79 & 3 & 36596 & 0.94 & 17401 & 25.34 & 9241 & \textbf{0.56} \\
ABC & 00000152 & 4 & 113009 & 0.75 & 4 & 189735 & 0.85 & 43241 & 1.8 & 9241 & \textbf{0.56} \\
ABC & 00000154 & 1 & 4193 & 2.62 & 1 & 4410 & 1.51 & 4801 & 39.46 & 4801 & \textbf{0.31} \\
ABC & 00000154 & 2 & 10113 & 0.91 & 2 & 10668 & 0.79 & 9241 & 7.4 & 9241 & \textbf{0.29} \\
ABC & 00000154 & 3 & 32801 & 0.55 & 3 & 36123 & 0.88 & 17401 & 7.32 & 17401 & \textbf{0.27} \\
ABC & 00000154 & 4 & 128737 & 0.61 & 4 & 130805 & 0.66 & 43241 & 4.25 & 43241 & \textbf{0.26} \\
ABC & 00000395 & 1 & 4209 & 13.47 & 1 & 3705 & 3.69 & 4801 & 4.61 & 4801 & \textbf{1.48} \\
ABC & 00000395 & 2 & 7185 & 2.61 & 2 & 6857 & 1.69 & 9241 & 2.11 & 9602 & \textbf{1.22} \\
ABC & 00000395 & 3 & 18305 & 1.14 & 3 & 19818 & 1.11 & 17401 & 27.34 & 14403 & \textbf{1.08} \\
ABC & 00000395 & 4 & 64625 & \textbf{0.77} & 4 & 77147 & 0.85 & 43241 & 1.41 & 27723 & 0.95\\
ABC & 00009942 & 1 & 3809 & 3.94 & 1 & 3769 & 1.83 & 4801 & 13.85 & 4834 & \textbf{0.49} \\
ABC & 00009942 & 2 & 7249 & 1.16 & 2 & 6931 & 1.13 & 9241 & 4.1 & 9602 & \textbf{0.39} \\
ABC & 00009942 & 3 & 20177 & 0.7 & 3 & 20774 & 0.81 & 17401 & 7.27 & 18482 & \textbf{0.36} \\
ABC & 00009942 & 4 & 66081 & 0.64 & 4 & 74478 & 0.7 & 43241 & 3.93 & 34802 & \textbf{0.34} \\
ABC & 00009943 & 1 & 3425 & 7.54 & 1 & 3870 & 2.11 & 4801 & 3.8 & 4801 & \textbf{0.36} \\
ABC & 00009943 & 2 & 6945 & 1.19 & 2 & 8112 & 1.06 & 9241 & 38.79 & 9241 & \textbf{0.34} \\
ABC & 00009943 & 3 & 21729 & 0.7 & 3 & 24827 & 0.85 & 17401 & 7.66 & 9241 & \textbf{0.34} \\
ABC & 00009943 & 4 & 85505 & 0.7 & 4 & 102868 & 0.7 & 43241 & 2.27 & 43241 & \textbf{0.34} \\
ABC & 00009947 & 1 & 3905 & 4.45 & 1 & 4398 & 1.34 & 4801 & 3.31 & 4801 & \textbf{0.33} \\
ABC & 00009947 & 2 & 8961 & 0.84 & 2 & 9030 & 0.76 & 9241 & 0.76 & 4801 & \textbf{0.33} \\
ABC & 00009947 & 3 & 32193 & 0.51 & 3 & 35498 & 0.67 & 17401 & 10.61 & 4801 & \textbf{0.33} \\
ABC & 00009947 & 4 & 101665 & 0.56 & 4 & 150771 & 0.59 & 43241 & 0.54 & 43241 & \textbf{0.29} \\
ABC & 00009949 & 1 & 4705 & 4.09 & 1 & 3963 & 1.84 & 4801 & 3.67 & 4801 & \textbf{0.98} \\
ABC & 00009949 & 2 & 8625 & 1.57 & 2 & 8368 & 1.04 & 9241 & 13.34 & 9602 & \textbf{0.77} \\
ABC & 00009949 & 3 & 24817 & 0.9 & 3 & 20858 & 0.8 & 17401 & 5.19 & 18482 & \textbf{0.67} \\
ABC & 00009949 & 4 & 91201 & 0.7 & 4 & 96127 & 0.74 & 43241 & 0.82 & 34802 & \textbf{0.64} \\
ABC & 00009951 & 1 & 3809 & 2.66 & 1 & 3917 & 1.5 & 4801 & 4.9 & 4801 & \textbf{1.07} \\
ABC & 00009951 & 2 & 7905 & 0.86 & 2 & 7488 & 0.66 & 9241 & 6.96 & 9241 & \textbf{0.26} \\
ABC & 00009951 & 3 & 23041 & 0.51 & 3 & 21252 & 0.58 & 17401 & 6.19 & 9241 & \textbf{0.26} \\
ABC & 00009951 & 4 & 75953 & 0.56 & 4 & 76991 & 0.54 & 43241 & 3.24 & 43241 & \textbf{0.25} \\
ABC & 00009955 & 1 & 4241 & 6.1 & 1 & 3808 & 2.32 & 4801 & 1.68 & 4801 & \textbf{0.73} \\
ABC & 00009955 & 2 & 7921 & 1.66 & 2 & 7379 & 1.37 & 9241 & 1.32 & 9241 & \textbf{0.6} \\
ABC & 00009955 & 3 & 21649 & 0.77 & 3 & 26796 & 0.96 & 17401 & 6.35 & 17401 & \textbf{0.58} \\
ABC & 00009955 & 4 & 74225 & 0.73 & 4 & 109203 & 0.76 & 43241 & 3.33 & 43241 & \textbf{0.56} \\
ABC & 00009956 & 1 & 4705 & 4.18 & 1 & 4180 & \textbf{2.97} & 4801 & 16.63 & 4851 & 4.5\\
ABC & 00009956 & 2 & 10209 & \textbf{1.35} & 2 & 11743 & 1.4 & 9241 & 14.2 & 4851 & 4.5\\
ABC & 00009956 & 3 & 29969 & 0.76 & 3 & 40593 & 0.99 & 17401 & 12.81 & 14403 & \textbf{0.54} \\
ABC & 00009956 & 4 & 146817 & 0.74 & 4 & 185697 & 0.77 & 43241 & 11.87 & 27723 & \textbf{0.5} \\
ABC & 00009959 & 1 & 4705 & 1.22 & 1 & 4770 & 1.08 & 4801 & 10.15 & 4801 & \textbf{0.35} \\
ABC & 00009959 & 2 & 9969 & 0.68 & 2 & 12148 & 0.71 & 9241 & 25.24 & 4801 & \textbf{0.35} \\
ABC & 00009959 & 3 & 38945 & 0.47 & 3 & 44176 & 0.61 & 17401 & 12.03 & 17401 & \textbf{0.33} \\
ABC & 00009959 & 4 & 153041 & 0.6 & 4 & 170038 & 0.61 & 43241 & 19.43 & 43241 & \textbf{0.3} \\
ABC & 00009963 & 1 & 3745 & 3.6 & 1 & 3909 & \textbf{0.89} & 4801 & 31.33 & 4801 & 0.98\\
ABC & 00009963 & 2 & 7793 & 0.81 & 2 & 7682 & 0.69 & 9241 & 2.63 & 9602 & \textbf{0.41} \\
ABC & 00009963 & 3 & 23217 & 0.68 & 3 & 22043 & 0.64 & 17401 & 1.44 & 9602 & \textbf{0.41} \\
ABC & 00009963 & 4 & 78545 & 0.74 & 4 & 89073 & 0.66 & 43241 & 3.49 & 34802 & \textbf{0.35} \\
ABC & 00009966 & 1 & 3617 & 5.65 & 1 & 3144 & 1.59 & 4801 & 0.51 & 4801 & \textbf{0.16} \\
ABC & 00009966 & 2 & 5217 & 1.02 & 2 & 4035 & 0.77 & 9241 & 1.13 & 9241 & \textbf{0.13} \\
ABC & 00009966 & 3 & 8401 & 0.68 & 3 & 6071 & 0.56 & 17401 & 1.91 & 17401 & \textbf{0.13} \\
ABC & 00009966 & 4 & 18337 & 0.54 & 4 & 10771 & 0.45 & 43241 & 1.32 & 43241 & \textbf{0.12} \\
ABC & 00009981 & 1 & 3425 & 4.48 & 1 & 3099 & 1.04 & 4801 & 3.34 & 4801 & \textbf{0.51} \\
ABC & 00009981 & 2 & 4737 & 1.63 & 2 & 3851 & 0.6 & 9241 & 2.95 & 4801 & \textbf{0.51} \\
ABC & 00009981 & 3 & 7201 & 0.95 & 3 & 6253 & \textbf{0.44} & 17401 & 3.5 & 4801 & 0.51\\
ABC & 00009981 & 4 & 15137 & 0.55 & 4 & 13958 & \textbf{0.46} & 43241 & 2.3 & 4801 & 0.51\\
ABC & 00009982 & 1 & 4705 & 1.94 & 1 & 4077 & 0.95 & 4801 & 1.63 & 4834 & \textbf{0.62} \\
ABC & 00009982 & 2 & 7281 & 1.59 & 2 & 8447 & 0.79 & 9241 & 5.03 & 9602 & \textbf{0.61} \\
ABC & 00009982 & 3 & 25393 & 0.7 & 3 & 23629 & 0.77 & 17401 & 0.61 & 18482 & \textbf{0.58} \\
ABC & 00009982 & 4 & 91217 & 0.68 & 4 & 98558 & 0.75 & 43241 & 3.54 & 18482 & \textbf{0.58} \\
ABC & 00009993 & 1 & 3425 & 2.83 & 1 & 3303 & 0.76 & 4801 & 2.45 & 4801 & \textbf{0.32} \\
ABC & 00009993 & 2 & 5601 & 0.94 & 2 & 4943 & 0.69 & 9241 & 0.68 & 9241 & \textbf{0.28} \\
ABC & 00009993 & 3 & 11329 & 0.51 & 3 & 12329 & 0.54 & 17401 & 0.59 & 17401 & \textbf{0.23} \\
ABC & 00009993 & 4 & 32609 & 0.64 & 4 & 33122 & 0.53 & 43241 & 0.54 & 17401 & \textbf{0.23} \\
ABC & 00009994 & 1 & 3905 & 2.64 & 1 & 3813 & 1.1 & 4801 & 5.98 & 4801 & \textbf{0.45} \\
ABC & 00009994 & 2 & 6897 & 1.11 & 2 & 6707 & 0.75 & 9241 & 4.94 & 9241 & \textbf{0.42} \\
ABC & 00009994 & 3 & 20737 & 0.63 & 3 & 22212 & 0.72 & 17401 & 3.01 & 17401 & \textbf{0.41} \\
ABC & 00009994 & 4 & 62881 & 0.75 & 4 & 90011 & 0.64 & 43241 & 0.84 & 43241 & \textbf{0.39} \\
ABC & 00009999 & 1 & 3905 & 2.37 & 1 & 4390 & 1.48 & 4801 & 0.67 & 4801 & \textbf{0.41} \\
ABC & 00009999 & 2 & 9873 & 0.61 & 2 & 10151 & 0.83 & 9241 & 1.06 & 9241 & \textbf{0.41} \\
ABC & 00009999 & 3 & 32801 & 0.53 & 3 & 35475 & 0.81 & 17401 & 0.67 & 9241 & \textbf{0.41} \\
ABC & 00009999 & 4 & 120689 & 0.55 & 4 & 168269 & 0.64 & 43241 & 0.64 & 9241 & \textbf{0.41} \\
Thingi10k (nglod32) & 252119 & 1 & 3233 & 19.14 & 1 & 3554 & 2.78 & 4801 & 3.09 & 4851 & \textbf{1.29} \\
Thingi10k (nglod32) & 252119 & 2 & 6241 & 3.62 & 2 & 5833 & 2.63 & 9241 & 2.16 & 9602 & \textbf{0.8} \\
Thingi10k (nglod32) & 252119 & 3 & 15681 & 1.18 & 3 & 14309 & 1.51 & 17401 & 1.87 & 18482 & \textbf{0.62} \\
Thingi10k (nglod32) & 252119 & 4 & 47777 & 0.73 & 4 & 51252 & 0.93 & 43241 & 1.11 & 34802 & \textbf{0.5} \\
Thingi10k (nglod32) & 313444 & 1 & 4113 & 5.8 & 1 & 4176 & 1.48 & 4801 & 8.44 & 4801 & \textbf{0.44} \\
Thingi10k (nglod32) & 313444 & 2 & 8993 & 1.48 & 2 & 9395 & 2.06 & 9241 & 8.57 & 9241 & \textbf{0.37} \\
Thingi10k (nglod32) & 313444 & 3 & 28097 & 0.64 & 3 & 29712 & 1.11 & 17401 & 10.26 & 17401 & \textbf{0.34} \\
Thingi10k (nglod32) & 313444 & 4 & 107201 & 0.56 & 4 & 105113 & 0.82 & 43241 & 3.58 & 43241 & \textbf{0.32} \\
Thingi10k (nglod32) & 316358 & 1 & 3953 & 7.84 & 1 & 4099 & 2.09 & 4801 & 14.8 & 4801 & \textbf{0.76} \\
Thingi10k (nglod32) & 316358 & 2 & 8369 & 1.97 & 2 & 9012 & 2.07 & 9241 & 6.41 & 9241 & \textbf{0.62} \\
Thingi10k (nglod32) & 316358 & 3 & 26145 & 0.72 & 3 & 27960 & 1.04 & 17401 & 5.65 & 18482 & \textbf{0.49} \\
Thingi10k (nglod32) & 316358 & 4 & 100417 & 0.58 & 4 & 103805 & 1.0 & 43241 & 6.68 & 34802 & \textbf{0.41} \\
Thingi10k (nglod32) & 354371 & 1 & 4001 & 6.65 & 1 & 4038 & 2.29 & 4801 & 6.61 & 4801 & \textbf{0.83} \\
Thingi10k (nglod32) & 354371 & 2 & 8529 & 2.3 & 2 & 8914 & 2.4 & 9241 & 6.11 & 9602 & \textbf{0.66} \\
Thingi10k (nglod32) & 354371 & 3 & 25521 & 0.79 & 3 & 28842 & 0.92 & 17401 & 5.44 & 18482 & \textbf{0.54} \\
Thingi10k (nglod32) & 354371 & 4 & 99009 & 0.58 & 4 & 107285 & 0.94 & 43241 & 2.85 & 34802 & \textbf{0.44} \\
Thingi10k (nglod32) & 398259 & 1 & 4321 & 5.18 & 1 & 5016 & 1.46 & 4801 & 18.34 & 4801 & \textbf{0.85} \\
Thingi10k (nglod32) & 398259 & 2 & 12161 & 1.2 & 2 & 13227 & 1.8 & 9241 & 8.46 & 9602 & \textbf{0.69} \\
Thingi10k (nglod32) & 398259 & 3 & 44769 & 0.7 & 3 & 45614 & 0.87 & 17401 & 19.45 & 18482 & \textbf{0.56} \\
Thingi10k (nglod32) & 398259 & 4 & 172017 & 0.56 & 4 & 176248 & 0.67 & 43241 & 5.0 & 34802 & \textbf{0.47} \\
Thingi10k (nglod32) & 441708 & 1 & 3985 & 8.55 & 1 & 4221 & 1.38 & 4801 & 9.71 & 4801 & \textbf{0.53} \\
Thingi10k (nglod32) & 441708 & 2 & 9473 & 1.3 & 2 & 9282 & 1.82 & 9241 & 13.36 & 9602 & \textbf{0.43} \\
Thingi10k (nglod32) & 441708 & 3 & 29249 & 0.64 & 3 & 29326 & 1.37 & 17401 & 7.82 & 18482 & \textbf{0.36} \\
Thingi10k (nglod32) & 441708 & 4 & 105745 & 0.51 & 4 & 108392 & 0.77 & 43241 & 10.08 & 34802 & \textbf{0.32} \\
Thingi10k (nglod32) & 44234 & 1 & 3617 & 15.59 & 1 & 3883 & 2.06 & 4801 & 8.47 & 4801 & \textbf{0.72} \\
Thingi10k (nglod32) & 44234 & 2 & 7217 & 1.95 & 2 & 7832 & 2.13 & 9241 & 6.02 & 9602 & \textbf{0.55} \\
Thingi10k (nglod32) & 44234 & 3 & 22049 & 0.72 & 3 & 23751 & 1.06 & 17401 & 6.78 & 18482 & \textbf{0.42} \\
Thingi10k (nglod32) & 44234 & 4 & 80657 & 0.55 & 4 & 86527 & 0.88 & 43241 & 3.2 & 34802 & \textbf{0.36} \\
Thingi10k (nglod32) & 47984 & 1 & 3377 & 13.37 & 1 & 3679 & 1.97 & 4801 & 4.46 & 4801 & \textbf{0.47} \\
Thingi10k (nglod32) & 47984 & 2 & 6977 & 1.98 & 2 & 6718 & 2.17 & 9241 & 4.45 & 9241 & \textbf{0.38} \\
Thingi10k (nglod32) & 47984 & 3 & 17953 & 0.75 & 3 & 18976 & 1.16 & 17401 & 7.68 & 17401 & \textbf{0.33} \\
Thingi10k (nglod32) & 47984 & 4 & 61857 & 0.62 & 4 & 66920 & 0.78 & 43241 & 2.03 & 43241 & \textbf{0.3} \\
Thingi10k (nglod32) & 527631 & 1 & 4145 & 8.25 & 1 & 3968 & 2.36 & 4801 & 8.14 & 4851 & \textbf{1.6} \\
Thingi10k (nglod32) & 527631 & 2 & 8353 & 2.35 & 2 & 8348 & 2.47 & 9241 & 4.9 & 9602 & \textbf{1.07} \\
Thingi10k (nglod32) & 527631 & 3 & 24369 & 1.18 & 3 & 27463 & 1.36 & 17401 & 18.42 & 18482 & \textbf{0.88} \\
Thingi10k (nglod32) & 527631 & 4 & 90337 & 0.71 & 4 & 103645 & 0.97 & 43241 & 5.35 & 27723 & \textbf{0.68} \\
Thingi10k (nglod32) & 53159 & 1 & 4001 & 10.45 & 1 & 4060 & 2.14 & 4801 & 17.57 & 4801 & \textbf{1.16} \\
Thingi10k (nglod32) & 53159 & 2 & 8337 & 1.9 & 2 & 8867 & 2.12 & 9241 & 5.51 & 9602 & \textbf{0.73} \\
Thingi10k (nglod32) & 53159 & 3 & 25665 & 0.74 & 3 & 28359 & 1.19 & 17401 & 6.48 & 18482 & \textbf{0.61} \\
Thingi10k (nglod32) & 53159 & 4 & 98369 & 0.58 & 4 & 106733 & 0.93 & 43241 & 4.18 & 27723 & \textbf{0.53} \\
Thingi10k (nglod32) & 58168 & 1 & 4001 & 11.38 & 1 & 3996 & \textbf{2.47} & 4801 & 14.38 & 4801 & 2.59\\
Thingi10k (nglod32) & 58168 & 2 & 7601 & 2.6 & 2 & 8674 & 2.38 & 9241 & 4.07 & 9602 & \textbf{2.13} \\
Thingi10k (nglod32) & 58168 & 3 & 24177 & \textbf{1.21} & 3 & 28481 & 1.41 & 17401 & 8.3 & 18482 & 1.98\\
Thingi10k (nglod32) & 58168 & 4 & 94833 & \textbf{0.91} & 4 & 108897 & 0.98 & 43241 & 2.94 & 34802 & 1.91\\
Thingi10k (nglod32) & 64444 & 1 & 3809 & 5.78 & 1 & 3695 & 1.34 & 4801 & 6.44 & 4801 & \textbf{0.59} \\
Thingi10k (nglod32) & 64444 & 2 & 7233 & 1.4 & 2 & 7315 & 1.67 & 9241 & 6.28 & 9241 & \textbf{0.52} \\
Thingi10k (nglod32) & 64444 & 3 & 19089 & 0.82 & 3 & 18956 & 1.78 & 17401 & 4.2 & 18482 & \textbf{0.45} \\
Thingi10k (nglod32) & 64444 & 4 & 72833 & 0.62 & 4 & 74023 & 1.01 & 43241 & 3.17 & 34802 & \textbf{0.4} \\
Thingi10k (nglod32) & 64764 & 1 & 3857 & 13.49 & 1 & 4030 & 2.2 & 4801 & 17.06 & 4851 & \textbf{1.45} \\
Thingi10k (nglod32) & 64764 & 2 & 8641 & 2.08 & 2 & 8701 & 2.57 & 9241 & 32.14 & 9602 & \textbf{0.78} \\
Thingi10k (nglod32) & 64764 & 3 & 25265 & 1.0 & 3 & 28076 & 1.39 & 17401 & 8.54 & 18482 & \textbf{0.6} \\
Thingi10k (nglod32) & 64764 & 4 & 95905 & 0.63 & 4 & 106399 & 0.88 & 43241 & 16.81 & 34802 & \textbf{0.51} \\
Thingi10k (nglod32) & 68380 & 1 & 3873 & 7.25 & 1 & 3936 & 2.39 & 4801 & 7.46 & 4801 & \textbf{0.81} \\
Thingi10k (nglod32) & 68380 & 2 & 7841 & 3.03 & 2 & 8059 & 2.25 & 9241 & 7.54 & 9241 & \textbf{0.64} \\
Thingi10k (nglod32) & 68380 & 3 & 23009 & 0.88 & 3 & 24148 & 1.24 & 17401 & 6.51 & 18482 & \textbf{0.52} \\
Thingi10k (nglod32) & 68380 & 4 & 84721 & 0.57 & 4 & 91139 & 0.99 & 43241 & 7.63 & 34802 & \textbf{0.43} \\
Thingi10k (nglod32) & 68381 & 1 & 3809 & 10.22 & 1 & 3902 & 2.57 & 4801 & 5.98 & 4851 & \textbf{1.54} \\
Thingi10k (nglod32) & 68381 & 2 & 7697 & 3.34 & 2 & 7938 & 2.36 & 9241 & 3.78 & 9602 & \textbf{0.92} \\
Thingi10k (nglod32) & 68381 & 3 & 22785 & 1.22 & 3 & 23535 & 1.05 & 17401 & 16.24 & 18482 & \textbf{0.7} \\
Thingi10k (nglod32) & 68381 & 4 & 83105 & 0.57 & 4 & 90934 & 0.91 & 43241 & 3.88 & 34802 & \textbf{0.57} \\
Thingi10k (nglod32) & 72870 & 1 & 3377 & 10.98 & 1 & 3755 & 2.06 & 4801 & 5.25 & 4801 & \textbf{0.88} \\
Thingi10k (nglod32) & 72870 & 2 & 5873 & 2.41 & 2 & 6897 & 3.13 & 9241 & 13.83 & 9602 & \textbf{0.71} \\
Thingi10k (nglod32) & 72870 & 3 & 18833 & 1.09 & 3 & 19076 & 1.29 & 17401 & 7.58 & 18482 & \textbf{0.57} \\
Thingi10k (nglod32) & 72870 & 4 & 64817 & 0.66 & 4 & 69142 & 1.24 & 43241 & 3.21 & 34802 & \textbf{0.48} \\
Thingi10k (nglod32) & 72960 & 1 & 4113 & 5.55 & 1 & 4058 & 1.86 & 4801 & 25.22 & 4801 & \textbf{0.74} \\
Thingi10k (nglod32) & 72960 & 2 & 8865 & 1.66 & 2 & 8809 & 2.41 & 9241 & 13.22 & 9602 & \textbf{0.57} \\
Thingi10k (nglod32) & 72960 & 3 & 26321 & 0.85 & 3 & 26675 & 1.08 & 17401 & 7.03 & 18482 & \textbf{0.45} \\
Thingi10k (nglod32) & 72960 & 4 & 97985 & 0.61 & 4 & 102136 & 0.94 & 43241 & 5.45 & 34802 & \textbf{0.39} \\
Thingi10k (nglod32) & 73075 & 1 & 3361 & 28.02 & 1 & 3445 & 1.96 & 4801 & 1.32 & 4801 & \textbf{0.38} \\
Thingi10k (nglod32) & 73075 & 2 & 5345 & 3.2 & 2 & 5445 & 1.86 & 9241 & 0.53 & 9241 & \textbf{0.31} \\
Thingi10k (nglod32) & 73075 & 3 & 13281 & 1.04 & 3 & 12996 & 1.09 & 17401 & 0.85 & 17401 & \textbf{0.28} \\
Thingi10k (nglod32) & 73075 & 4 & 41057 & 0.59 & 4 & 43650 & 0.74 & 43241 & 5.38 & 43241 & \textbf{0.24} \\
Thingi10k (nglod32) & 75496 & 1 & 3713 & 6.92 & 1 & 4061 & 1.42 & 4801 & 10.48 & 4801 & \textbf{0.52} \\
Thingi10k (nglod32) & 75496 & 2 & 7937 & 1.61 & 2 & 8848 & 2.24 & 9241 & 8.35 & 9241 & \textbf{0.42} \\
Thingi10k (nglod32) & 75496 & 3 & 25233 & 0.67 & 3 & 27317 & 0.9 & 17401 & 7.04 & 17401 & \textbf{0.37} \\
Thingi10k (nglod32) & 75496 & 4 & 97489 & 0.51 & 4 & 100382 & 0.7 & 43241 & 4.79 & 43241 & \textbf{0.32} \\
Thingi10k (nglod32) & 75655 & 1 & 3793 & 9.29 & 1 & 3712 & 1.55 & 4801 & 10.97 & 4801 & \textbf{0.52} \\
Thingi10k (nglod32) & 75655 & 2 & 7537 & 1.91 & 2 & 7121 & 2.43 & 9241 & 4.17 & 9241 & \textbf{0.41} \\
Thingi10k (nglod32) & 75655 & 3 & 19441 & 0.63 & 3 & 20238 & 1.01 & 17401 & 5.01 & 18482 & \textbf{0.33} \\
Thingi10k (nglod32) & 75655 & 4 & 69585 & 0.55 & 4 & 72827 & 0.83 & 43241 & 3.84 & 34802 & \textbf{0.3} \\
Thingi10k (nglod32) & 75656 & 1 & 3857 & 6.35 & 1 & 3826 & 1.36 & 4801 & 9.71 & 4801 & \textbf{0.42} \\
Thingi10k (nglod32) & 75656 & 2 & 7873 & 1.6 & 2 & 7669 & 1.72 & 9241 & 9.74 & 9241 & \textbf{0.35} \\
Thingi10k (nglod32) & 75656 & 3 & 21665 & 0.63 & 3 & 22145 & 1.14 & 17401 & 9.18 & 18482 & \textbf{0.31} \\
Thingi10k (nglod32) & 75656 & 4 & 78673 & 0.56 & 4 & 79776 & 0.84 & 43241 & 4.43 & 34802 & \textbf{0.28} \\
Thingi10k (nglod32) & 75662 & 1 & 3889 & 8.27 & 1 & 3833 & 2.02 & 4801 & 8.84 & 4851 & \textbf{1.04} \\
Thingi10k (nglod32) & 75662 & 2 & 8081 & 2.12 & 2 & 7862 & 2.19 & 9241 & 5.86 & 9602 & \textbf{0.63} \\
Thingi10k (nglod32) & 75662 & 3 & 21553 & 1.01 & 3 & 23729 & 1.02 & 17401 & 5.97 & 18482 & \textbf{0.49} \\
Thingi10k (nglod32) & 75662 & 4 & 81649 & 0.61 & 4 & 88435 & 0.91 & 43241 & 5.16 & 34802 & \textbf{0.42} \\
Thingi10k (nglod32) & 75665 & 1 & 3969 & 6.53 & 1 & 4074 & 1.61 & 4801 & 36.74 & 4801 & \textbf{0.57} \\
Thingi10k (nglod32) & 75665 & 2 & 7905 & 1.83 & 2 & 8765 & 1.85 & 9241 & 8.28 & 9602 & \textbf{0.5} \\
Thingi10k (nglod32) & 75665 & 3 & 25793 & 0.66 & 3 & 27044 & 1.13 & 17401 & 4.54 & 18482 & \textbf{0.4} \\
Thingi10k (nglod32) & 75665 & 4 & 96465 & 0.52 & 4 & 95032 & 0.84 & 43241 & 7.43 & 34802 & \textbf{0.35} \\
Thingi10k (nglod32) & 76277 & 1 & 3857 & 9.5 & 1 & 3905 & 1.83 & 4801 & 2.13 & 4851 & \textbf{1.14} \\
Thingi10k (nglod32) & 76277 & 2 & 7905 & 2.15 & 2 & 7746 & 1.83 & 9241 & 3.19 & 9602 & \textbf{0.72} \\
Thingi10k (nglod32) & 76277 & 3 & 22769 & 0.79 & 3 & 23927 & 1.13 & 17401 & 3.53 & 18482 & \textbf{0.58} \\
Thingi10k (nglod32) & 76277 & 4 & 79569 & 0.56 & 4 & 96689 & 0.85 & 43241 & 1.54 & 34802 & \textbf{0.49} \\
Thingi10k (nglod32) & 77245 & 1 & 3809 & 5.54 & 1 & 4057 & 1.31 & 4801 & 10.71 & 4801 & \textbf{0.85} \\
Thingi10k (nglod32) & 77245 & 2 & 9025 & 1.37 & 2 & 8872 & 1.87 & 9241 & 8.84 & 9602 & \textbf{0.59} \\
Thingi10k (nglod32) & 77245 & 3 & 25393 & 0.69 & 3 & 26758 & 1.13 & 17401 & 6.26 & 18482 & \textbf{0.48} \\
Thingi10k (nglod32) & 77245 & 4 & 98017 & 0.64 & 4 & 99969 & 0.89 & 43241 & 5.42 & 34802 & \textbf{0.42} \\
Thingi10k (nglod32) & 78671 & 1 & 3041 & 12.46 & 1 & 3497 & 2.59 & 4801 & 2.44 & 4801 & \textbf{0.55} \\
Thingi10k (nglod32) & 78671 & 2 & 5921 & 2.68 & 2 & 5800 & 2.31 & 9241 & 2.64 & 9241 & \textbf{0.44} \\
Thingi10k (nglod32) & 78671 & 3 & 14417 & 0.91 & 3 & 14401 & 1.33 & 17401 & 4.12 & 18482 & \textbf{0.35} \\
Thingi10k (nglod32) & 78671 & 4 & 46785 & 0.57 & 4 & 50147 & 1.03 & 43241 & 0.85 & 34802 & \textbf{0.3} \\
Thingi10k (nglod32) & 79241 & 1 & 3873 & 5.31 & 1 & 4049 & 1.77 & 4801 & 9.17 & 4801 & \textbf{0.73} \\
Thingi10k (nglod32) & 79241 & 2 & 7873 & 2.47 & 2 & 8644 & 2.0 & 9241 & 5.51 & 9241 & \textbf{0.59} \\
Thingi10k (nglod32) & 79241 & 3 & 24801 & 0.93 & 3 & 26944 & 1.19 & 17401 & 34.16 & 17401 & \textbf{0.5} \\
Thingi10k (nglod32) & 79241 & 4 & 94193 & 0.67 & 4 & 100094 & 0.92 & 43241 & 5.33 & 43241 & \textbf{0.44} \\
Thingi10k (nglod32) & 90889 & 1 & 3425 & 11.27 & 1 & 3940 & 1.9 & 4801 & 7.05 & 4801 & \textbf{0.83} \\
Thingi10k (nglod32) & 90889 & 2 & 7153 & 1.99 & 2 & 8009 & 2.99 & 9241 & 4.86 & 9602 & \textbf{0.63} \\
Thingi10k (nglod32) & 90889 & 3 & 22961 & 0.83 & 3 & 25571 & 1.19 & 17401 & 7.0 & 18482 & \textbf{0.49} \\
Thingi10k (nglod32) & 90889 & 4 & 83729 & 0.69 & 4 & 90755 & 0.89 & 43241 & 4.97 & 34802 & \textbf{0.42} \\
Thingi10k (nglod32) & 92763 & 1 & 3521 & 10.55 & 1 & 3651 & 2.04 & 4801 & 5.98 & 4801 & \textbf{0.71} \\
Thingi10k (nglod32) & 92763 & 2 & 7057 & 2.36 & 2 & 7016 & 2.2 & 9241 & 5.18 & 9241 & \textbf{0.55} \\
Thingi10k (nglod32) & 92763 & 3 & 17313 & 0.87 & 3 & 19578 & 1.29 & 17401 & 6.99 & 17401 & \textbf{0.45} \\
Thingi10k (nglod32) & 92763 & 4 & 67137 & 0.63 & 4 & 66254 & 0.86 & 43241 & 3.4 & 43241 & \textbf{0.41} \\
Thingi10k (nglod32) & 92880 & 1 & 3425 & 11.25 & 1 & 3711 & 1.99 & 4801 & 2.32 & 4801 & \textbf{0.6} \\
Thingi10k (nglod32) & 92880 & 2 & 7041 & 2.52 & 2 & 6919 & 1.8 & 9241 & 1.07 & 9241 & \textbf{0.49} \\
Thingi10k (nglod32) & 92880 & 3 & 18961 & 0.83 & 3 & 19530 & 1.06 & 17401 & 4.59 & 17401 & \textbf{0.42} \\
Thingi10k (nglod32) & 92880 & 4 & 65841 & 0.58 & 4 & 71273 & 0.78 & 43241 & 0.7 & 34802 & \textbf{0.35} \\
Thingi10k (nglod32) & 95444 & 1 & 3521 & 17.75 & 1 & 3675 & 2.58 & 4801 & 5.91 & 4801 & \textbf{1.3} \\
Thingi10k (nglod32) & 95444 & 2 & 6225 & 2.42 & 2 & 6715 & 2.22 & 9241 & 4.7 & 9602 & \textbf{0.84} \\
Thingi10k (nglod32) & 95444 & 3 & 17457 & 0.92 & 3 & 18454 & 1.4 & 17401 & 4.21 & 18482 & \textbf{0.68} \\
Thingi10k (nglod32) & 95444 & 4 & 61793 & 0.6 & 4 & 68948 & 0.91 & 43241 & 2.13 & 27723 & \textbf{0.56} \\
Thingi10k (nglod32) & 96481 & 1 & 3361 & 12.9 & 1 & 3395 & 1.5 & 4801 & 0.6 & 4801 & \textbf{0.48} \\
Thingi10k (nglod32) & 96481 & 2 & 5393 & 2.77 & 2 & 5116 & 1.97 & 9241 & 1.74 & 9241 & \textbf{0.4} \\
Thingi10k (nglod32) & 96481 & 3 & 12305 & 1.32 & 3 & 12140 & 1.46 & 17401 & 1.58 & 18482 & \textbf{0.37} \\
Thingi10k (nglod32) & 96481 & 4 & 35553 & 0.61 & 4 & 39778 & 2.22 & 43241 & 2.42 & 27723 & \textbf{0.33} \\
Thingi10k & SapphosHead & 1 & 3729 & 4.85 & 1 & 3735 & 8.61 & 4801 & 4.09 & 4801 & \textbf{0.48} \\
Thingi10k & SapphosHead & 2 & 7137 & 1.4 & 2 & 7030 & 2.39 & 9241 & 43.15 & 9241 & \textbf{0.38} \\
Thingi10k & SapphosHead & 3 & 19361 & 0.6 & 3 & 19569 & 1.39 & 17401 & 2.48 & 17401 & \textbf{0.34} \\
Thingi10k & SapphosHead & 4 & 67761 & 0.5 & 4 & 69150 & 0.88 & 43241 & 17.91 & 43241 & \textbf{0.31} \\
Thingi10k & cat & 1 & 4241 & 2.8 & 1 & 3867 & 7.22 & 4801 & 6.18 & 4801 & \textbf{0.48} \\
Thingi10k & cat & 2 & 8097 & 0.81 & 2 & 7548 & 2.33 & 9241 & 6.75 & 9602 & \textbf{0.38} \\
Thingi10k & cat & 3 & 22081 & 0.45 & 3 & 21148 & 1.06 & 17401 & 4.27 & 18482 & \textbf{0.31} \\
Thingi10k & cat & 4 & 76865 & 0.45 & 4 & 74654 & 0.8 & 43241 & 11.88 & 34802 & \textbf{0.29} \\
Thingi10k & dea & 1 & 4193 & 2.28 & 1 & 4062 & 4.92 & 4801 & 37.21 & 4801 & \textbf{0.39} \\
Thingi10k & dea & 2 & 8945 & 0.86 & 2 & 8709 & 1.73 & 9241 & 7.69 & 9241 & \textbf{0.34} \\
Thingi10k & dea & 3 & 25809 & 0.52 & 3 & 26569 & 1.17 & 17401 & 18.47 & 17401 & \textbf{0.31} \\
Thingi10k & dea & 4 & 95745 & 0.51 & 4 & 96854 & 0.83 & 43241 & 35.31 & 43241 & \textbf{0.3} \\
Thingi10k & ghost & 1 & 4593 & 1.33 & 1 & 4565 & 3.88 & 4801 & 15.01 & 4834 & \textbf{0.49} \\
Thingi10k & ghost & 2 & 10865 & 0.56 & 2 & 11187 & 1.64 & 9241 & 18.5 & 9602 & \textbf{0.4} \\
Thingi10k & ghost & 3 & 36177 & 0.5 & 3 & 36602 & 1.08 & 17401 & 10.03 & 18482 & \textbf{0.36} \\
Thingi10k & ghost & 4 & 137649 & 0.45 & 4 & 136566 & 0.87 & 43241 & 2.82 & 34802 & \textbf{0.34} \\
Thingi10k & goyle & 1 & 4641 & \textbf{3.53} & 1 & 4820 & 4.65 & 4801 & 9.92 & 4851 & 3.78\\
Thingi10k & goyle & 2 & 11825 & \textbf{1.54} & 2 & 13268 & 1.9 & 9241 & 10.3 & 4851 & 3.78\\
Thingi10k & goyle & 3 & 40897 & \textbf{0.8} & 3 & 46922 & 1.31 & 17401 & 14.16 & 14403 & 3.04\\
Thingi10k & goyle & 4 & 172081 & \textbf{0.59} & 4 & 179415 & 0.88 & 43241 & 8.68 & 27723 & 2.84\\
Thingi10k & lincoln & 1 & 4321 & 3.35 & 1 & 3917 & 7.32 & 4801 & 20.16 & 4801 & \textbf{0.57} \\
Thingi10k & lincoln & 2 & 8641 & 1.16 & 2 & 8041 & 2.2 & 9241 & 6.0 & 9241 & \textbf{0.47} \\
Thingi10k & lincoln & 3 & 23441 & 0.58 & 3 & 24629 & 1.21 & 17401 & 6.92 & 18482 & \textbf{0.41} \\
Thingi10k & lincoln & 4 & 84977 & 0.59 & 4 & 90383 & 0.9 & 43241 & 2.96 & 34802 & \textbf{0.37} \\
Thingi10k & pensatore & 1 & 4705 & 2.33 & 1 & 5246 & 3.31 & 4801 & 8.22 & 4801 & \textbf{0.7} \\
Thingi10k & pensatore & 2 & 13297 & 0.76 & 2 & 14643 & 1.66 & 9241 & 12.1 & 9241 & \textbf{0.56} \\
Thingi10k & pensatore & 3 & 49649 & \textbf{0.46} & 3 & 51595 & 1.15 & 17401 & 14.26 & 18482 & 0.46\\
Thingi10k & pensatore & 4 & 195457 & 0.4 & 4 & 197829 & 0.73 & 43241 & 6.04 & 34802 & \textbf{0.39} \\
Thingi10k & chinese & 1 & 4401 & 5.49 & 1 & 4161 & 6.19 & 4801 & 7.49 & 4851 & \textbf{1.69} \\
Thingi10k & chinese & 2 & 9409 & 2.25 & 2 & 9301 & 2.54 & 9241 & 5.6 & 9602 & \textbf{0.99} \\
Thingi10k & chinese & 3 & 28097 & \textbf{0.76} & 3 & 30119 & 1.28 & 17401 & 24.87 & 18482 & 0.78\\
Thingi10k & chinese & 4 & 106209 & \textbf{0.59} & 4 & 115352 & 0.88 & 43241 & 16.64 & 34802 & 0.65\\
DHFSlicer & airplane & 1 & 3713 & 3.86 & 1 & 3370 & 1.2 & 4801 & 8.45 & 4801 & \textbf{0.19} \\
DHFSlicer & airplane & 2 & 5889 & 0.94 & 2 & 4989 & 1.4 & 9241 & 0.46 & 9241 & \textbf{0.17} \\
DHFSlicer & airplane & 3 & 12321 & 0.45 & 3 & 10869 & 1.22 & 17401 & 0.56 & 17401 & \textbf{0.17} \\
DHFSlicer & airplane & 4 & 33937 & 0.5 & 4 & 34288 & 1.05 & 43241 & 0.98 & 43241 & \textbf{0.16} \\
DHFSlicer & batman & 1 & 4049 & 3.83 & 1 & 4118 & 1.12 & 4801 & 51.13 & 4801 & \textbf{0.3} \\
DHFSlicer & batman & 2 & 8193 & 0.87 & 2 & 8806 & 1.47 & 9241 & 6.64 & 9241 & \textbf{0.26} \\
DHFSlicer & batman & 3 & 26273 & 0.52 & 3 & 27077 & 0.85 & 17401 & 5.96 & 17401 & \textbf{0.25} \\
DHFSlicer & batman & 4 & 96913 & 0.51 & 4 & 97837 & 0.82 & 43241 & 5.46 & 43241 & \textbf{0.24} \\
DHFSlicer & bimba & 1 & 3745 & 8.03 & 1 & 4053 & 1.47 & 4801 & 9.12 & 4801 & \textbf{0.52} \\
DHFSlicer & bimba & 2 & 7713 & 1.71 & 2 & 8638 & 2.11 & 9241 & 10.71 & 9241 & \textbf{0.43} \\
DHFSlicer & bimba & 3 & 25217 & 0.67 & 3 & 26168 & 1.32 & 17401 & 9.79 & 17401 & \textbf{0.39} \\
DHFSlicer & bimba & 4 & 94561 & 0.61 & 4 & 93812 & 0.98 & 43241 & 5.62 & 34802 & \textbf{0.35} \\
DHFSlicer & bu & 1 & 3649 & 9.47 & 1 & 3788 & 1.63 & 4801 & 7.58 & 4801 & \textbf{0.44} \\
DHFSlicer & bu & 2 & 7153 & 1.56 & 2 & 7422 & 2.0 & 9241 & 6.93 & 9241 & \textbf{0.36} \\
DHFSlicer & bu & 3 & 21969 & 0.62 & 3 & 22378 & 1.12 & 17401 & 5.79 & 18482 & \textbf{0.3} \\
DHFSlicer & bu & 4 & 82801 & 0.52 & 4 & 86162 & 0.84 & 43241 & 6.13 & 34802 & \textbf{0.27} \\
DHFSlicer & buddha & 1 & 4017 & 5.59 & 1 & 3788 & 1.99 & 4801 & 56.91 & 4801 & \textbf{0.71} \\
DHFSlicer & buddha & 2 & 9201 & 2.33 & 2 & 7422 & 2.04 & 9241 & 31.3 & 9241 & \textbf{0.58} \\
DHFSlicer & buddha & 3 & 27473 & 0.99 & 3 & 22378 & 1.18 & 17401 & 14.35 & 18482 & \textbf{0.49} \\
DHFSlicer & buddha & 4 & 105713 & 0.62 & 4 & 86162 & 0.89 & 43241 & 6.32 & 34802 & \textbf{0.42} \\
DHFSlicer & bumpy & 1 & 4321 & 8.77 & 1 & 5497 & 2.39 & 4801 & 11.3 & 4851 & \textbf{1.18} \\
DHFSlicer & bumpy & 2 & 13777 & 1.0 & 2 & 16450 & 1.47 & 9241 & 19.41 & 9602 & \textbf{0.72} \\
DHFSlicer & bumpy & 3 & 54065 & \textbf{0.47} & 3 & 61484 & 0.75 & 17401 & 15.88 & 18482 & 0.56\\
DHFSlicer & bumpy & 4 & 224977 & \textbf{0.43} & 4 & 240875 & 0.65 & 43241 & 9.07 & 34802 & 0.43\\
DHFSlicer & bunnies & 1 & 4161 & 5.99 & 1 & 4164 & 1.85 & 4801 & 22.93 & 4801 & \textbf{0.59} \\
DHFSlicer & bunnies & 2 & 9025 & 1.59 & 2 & 9344 & 2.14 & 9241 & 17.68 & 9241 & \textbf{0.48} \\
DHFSlicer & bunnies & 3 & 27297 & 0.63 & 3 & 29137 & 1.08 & 17401 & 12.5 & 18482 & \textbf{0.41} \\
DHFSlicer & bunnies & 4 & 105873 & 0.53 & 4 & 107809 & 0.84 & 43241 & 4.81 & 34802 & \textbf{0.35} \\
DHFSlicer & chair & 1 & 4001 & 6.22 & 1 & 3637 & 2.09 & 4801 & 1.54 & 4834 & \textbf{0.68} \\
DHFSlicer & chair & 2 & 6305 & 1.61 & 2 & 6071 & 2.03 & 9241 & 0.88 & 9602 & \textbf{0.5} \\
DHFSlicer & chair & 3 & 17265 & 0.72 & 3 & 15404 & 1.15 & 17401 & 7.27 & 18482 & \textbf{0.38} \\
DHFSlicer & chair & 4 & 52017 & 0.55 & 4 & 53575 & 1.1 & 43241 & 0.54 & 34802 & \textbf{0.32} \\
DHFSlicer & david & 1 & 4241 & 6.79 & 1 & 3453 & 2.03 & 4801 & 10.54 & 4801 & \textbf{0.99} \\
DHFSlicer & david & 2 & 10945 & 1.96 & 2 & 5590 & 1.84 & 9241 & 23.23 & 9602 & \textbf{0.73} \\
DHFSlicer & david & 3 & 35553 & 0.74 & 3 & 13547 & 1.02 & 17401 & 10.32 & 18482 & \textbf{0.56} \\
DHFSlicer & david & 4 & 134865 & 0.55 & 4 & 47053 & 0.89 & 43241 & 7.86 & 34802 & \textbf{0.46} \\
DHFSlicer & feline & 1 & 3889 & 11.77 & 1 & 3843 & 2.26 & 4801 & 4.72 & 3234 & \textbf{1.13} \\
DHFSlicer & feline & 2 & 8465 & 2.4 & 2 & 7409 & 2.18 & 9241 & 3.56 & 9602 & \textbf{0.62} \\
DHFSlicer & feline & 3 & 22273 & 0.8 & 3 & 21368 & 0.98 & 17401 & 4.1 & 18482 & \textbf{0.47} \\
DHFSlicer & feline & 4 & 74945 & 0.51 & 4 & 78790 & 0.8 & 43241 & 2.77 & 34802 & \textbf{0.38} \\
DHFSlicer & fertility & 1 & 3473 & 14.41 & 1 & 3800 & 1.71 & 4801 & 15.15 & 4801 & \textbf{0.57} \\
DHFSlicer & fertility & 2 & 7121 & 1.39 & 2 & 7378 & 1.94 & 9241 & 2.25 & 9602 & \textbf{0.4} \\
DHFSlicer & fertility & 3 & 20417 & 0.7 & 3 & 22018 & 1.19 & 17401 & 3.56 & 18482 & \textbf{0.35} \\
DHFSlicer & fertility & 4 & 73441 & 0.55 & 4 & 79928 & 0.76 & 43241 & 2.15 & 34802 & \textbf{0.32} \\
DHFSlicer & lionleft & 1 & 4113 & 5.64 & 1 & 4059 & 1.55 & 4801 & 9.45 & 4801 & \textbf{0.72} \\
DHFSlicer & lionleft & 2 & 8865 & 1.67 & 2 & 8820 & 2.45 & 9241 & 19.91 & 9602 & \textbf{0.54} \\
DHFSlicer & lionleft & 3 & 26321 & 0.84 & 3 & 26674 & 1.58 & 17401 & 9.64 & 18482 & \textbf{0.43} \\
DHFSlicer & lionleft & 4 & 98193 & 0.64 & 4 & 101843 & 0.88 & 43241 & 5.78 & 34802 & \textbf{0.38} \\
DHFSlicer & maxplank & 1 & 4081 & 3.88 & 1 & 4357 & 1.05 & 4801 & 33.02 & 4801 & \textbf{0.36} \\
DHFSlicer & maxplank & 2 & 9921 & 0.9 & 2 & 10238 & 1.56 & 9241 & 32.27 & 9241 & \textbf{0.31} \\
DHFSlicer & maxplank & 3 & 31761 & 0.49 & 3 & 32924 & 0.92 & 17401 & 13.51 & 18482 & \textbf{0.29} \\
DHFSlicer & maxplank & 4 & 121361 & 0.48 & 4 & 121565 & 0.78 & 43241 & 6.57 & 34802 & \textbf{0.28} \\
DHFSlicer & moai & 1 & 4145 & 2.35 & 1 & 4011 & 5.38 & 4801 & 9.35 & 4801 & \textbf{0.3} \\
DHFSlicer & moai & 2 & 6961 & 0.78 & 2 & 8171 & 1.81 & 9241 & 4.93 & 9241 & \textbf{0.3} \\
DHFSlicer & moai & 3 & 23921 & 0.75 & 3 & 23935 & 1.14 & 17401 & 7.38 & 17401 & \textbf{0.29} \\
DHFSlicer & moai & 4 & 86321 & 0.55 & 4 & 85978 & 1.0 & 43241 & 2.43 & 17401 & \textbf{0.29} \\
DHFSlicer & rockerarm & 1 & 3473 & 26.28 & 1 & 3723 & 2.06 & 4801 & 2.25 & 4801 & \textbf{0.36} \\
DHFSlicer & rockerarm & 2 & 7121 & 1.96 & 2 & 7125 & 2.28 & 9241 & 1.0 & 9241 & \textbf{0.33} \\
DHFSlicer & rockerarm & 3 & 18865 & 0.83 & 3 & 21723 & 1.17 & 17401 & 4.22 & 17401 & \textbf{0.31} \\
DHFSlicer & rockerarm & 4 & 68497 & 0.61 & 4 & 77485 & 0.9 & 43241 & 2.08 & 43241 & \textbf{0.3} \\
DHFSlicer & dragon & 1 & 4401 & 6.17 & 1 & 3979 & 6.29 & 4801 & 10.66 & 4801 & \textbf{1.58} \\
DHFSlicer & dragon & 2 & 9185 & 1.97 & 2 & 8787 & 2.38 & 9241 & 20.75 & 9602 & \textbf{0.97} \\
DHFSlicer & dragon & 3 & 24593 & 0.77 & 3 & 28747 & 1.29 & 17401 & 13.33 & 18482 & \textbf{0.74} \\
DHFSlicer & dragon & 4 & 97377 & \textbf{0.54} & 4 & 110409 & 0.91 & 43241 & 26.25 & 34802 & 0.62\\
DHFSlicer & vase & 1 & 4513 & 3.44 & 1 & 4436 & 4.45 & 4801 & 8.68 & 4801 & \textbf{1.39} \\
DHFSlicer & vase & 2 & 10353 & 1.58 & 2 & 10880 & 2.4 & 9241 & 6.97 & 9602 & \textbf{0.99} \\
DHFSlicer & vase & 3 & 33473 & 1.14 & 3 & 36986 & 1.48 & 17401 & 28.69 & 18482 & \textbf{0.77} \\
DHFSlicer & vase & 4 & 132193 & \textbf{0.57} & 4 & 143116 & 1.01 & 43241 & 5.03 & 34802 & 0.63\\
Stanford & david700kf & 1 & 3425 & 8.57 & 1 & 3453 & 8.49 & 4801 & 3.12 & 4801 & \textbf{0.76} \\
Stanford & david700kf & 2 & 5985 & 2.3 & 2 & 5590 & 3.08 & 9241 & 12.29 & 9602 & \textbf{0.56} \\
Stanford & david700kf & 3 & 13409 & 1.04 & 3 & 13547 & 1.44 & 17401 & 5.0 & 18482 & \textbf{0.45} \\
Stanford & david700kf & 4 & 43457 & 0.65 & 4 & 47053 & 0.9 & 43241 & 0.84 & 34802 & \textbf{0.39} \\
Stanford & happy & 1 & 4369 & 6.85 & 1 & 3788 & 8.0 & 4801 & 34.03 & 4801 & \textbf{1.95} \\
Stanford & happy & 2 & 9569 & 2.57 & 2 & 7422 & 3.38 & 9241 & 3.32 & 9602 & \textbf{1.0} \\
Stanford & happy & 3 & 27857 & 1.03 & 3 & 22378 & 1.52 & 17401 & 3.65 & 18482 & \textbf{0.82} \\
Stanford & happy & 4 & 73825 & 0.71 & 4 & 86162 & 1.03 & 43241 & 3.81 & 27723 & \textbf{0.65} \\
Stanford & xyzrgb & 1 & 4001 & 5.38 & 1 & 3633 & 6.38 & 4801 & 5.34 & 4801 & \textbf{1.68} \\
Stanford & xyzrgb & 2 & 7249 & 2.94 & 2 & 6564 & 3.0 & 9241 & 2.68 & 9602 & \textbf{1.07} \\
Stanford & xyzrgb & 3 & 17505 & 1.19 & 3 & 18652 & 1.61 & 17401 & 3.25 & 18482 & \textbf{0.84} \\
Stanford & xyzrgb & 4 & 60257 & 0.72 & 4 & 71175 & 0.95 & 43241 & 0.84 & 34802 & \textbf{0.69} \\
    \end{longtable}
}

\twocolumn

\end{document}